# Granular Framework for Construction Material Price Forecasting: Econometric and Machine-Learning Approaches

Boge Lyu[1]; Qianye Yin[2]; Iris Denise Tommelein, Ph.D., M.ASCE[3]; Hanyang Liu[4]; Karnamohit Ranka, Ph.D.[5]; Karthik Yeluripati[6]; and Junzhe Shi, Ph.D.[7]

**Abstract:** The persistent volatility of construction material prices poses significant risks to cost estimation, budgeting, and project delivery, underscoring the urgent need for granular and scalable forecasting methods. This study develops a forecasting framework that leverages the Construction Specifications Institute (CSI) MasterFormat as the target data structure, enabling predictions at the six-digit Section level and supporting detailed cost projections across a wide spectrum of building materials. To enhance predictive accuracy, the framework integrates explanatory variables such as raw material prices, commodity indexes, and macroeconomic indicators. Five time-series models, long short-term memory (LSTM), autoregressive integrated moving average (ARIMA), vector error correction model (VECM), vanilla Transformer, and Chronos-Bolt, were evaluated under both baseline configurations (using CSI data only) and extended versions with explanatory variables. Results demonstrate that incorporating explanatory variables significantly improves predictive performance across all models. Among the tested approaches, the LSTM model consistently achieved the highest accuracy, with root mean squared error (RMSE) values as low as 4.19 and mean absolute percentage error (MAPE) values of 1.60 based on Division 06 as a representative demonstration, representing improvements of up to 59% over the traditional statistical time-series model, ARIMA. Validation across multiple CSI Divisions within the residential-material confirmed the framework's scalability, while Division 06 (Wood, Plastics, and Composites) is presented in detail as a demonstration case. This research offers a robust methodology that enables owners and contractors to improve budgeting practices and achieve more reliable cost estimation at the *definitive* level.  

**Practical Applications:** Construction material prices can change quickly, making it difficult for developers, owners, contractors, subcontractors, estimators, suppliers, and procurement teams to prepare reliable budgets, bids, and purchasing plans. This study presents an artificial intelligence (AI)-based forecasting approach that can accurately predict material prices in the coming months and quarters by combining historical cost information with market signals, such as commodity price changes and broader economic conditions. The approach addresses a practical gap in construction cost planning, where project teams still lack a widely adopted and consistent way to forecast future material prices for budgeting and procurement decisions. In practice, this approach can help teams update material unit costs, check escalation allowances, identify materials exposed to future price increases, and decide when to purchase or lock in pricing during preconstruction. For developers and owners, the forecasts can also support early development planning and pro forma analysis by improving hard-cost assumptions and testing cost-escalation scenarios. The approach provides a useful reference point for budgeting, cost review, procurement planning, and risk management.

## Introduction

### *Motivation*

The construction industry continues to demonstrate steady long-term growth, with global activity projected to reach US$9.8 trillion by 2026 (Shaista 2025). Major upcoming programs in the United States, such as the Los Angeles 2028 Olympics and Taiwan Semiconductor Manufacturing Company's (TSMC) fabrication facility in Arizona (Budzier and Flyvbjerg 2024; Roccasecca et al. 2025), highlight the scale of high-value projects in the near future. However, volatility in construction material prices has emerged as a critical challenge, creating significant uncertainty for contractors in project planning,

[1]Data Scientist, Quotr AI, Castro Valley, CA 94546. Email: boge_lyu@quotr.ai

[2]Researcher, Bert W. Wasserman Dept. of Economics and Finance, Baruch College, City Univ. of New York, New York, NY 10010. Email: qianye.yin@baruch.cuny.edu

[3]Professor, Dept. of Civil and Environmental Engineering, Univ. of California, Berkeley, 760 Davis Hall, Berkeley, CA 94720. ORCID: https://orcid.org/0000-0002-9941-6596. Email: p2sl@berkeley.edu

[4]CEO, Quotr AI, Castro Valley, CA 94546. Email: liu@quotr.ai

[5]Researcher, Lawrence Berkeley National Lab, Univ. of California, Berkeley, CA 94720. Email: kranka@lbl.gov

[6]AI Engineer, Quotr AI, Castro Valley, CA 94546. Email: karthik@quotr.ai

[7]CTO, Quotr AI, Castro Valley, CA 94546 (corresponding author). ORCID: https://orcid.org/0000-0003-2337-231X. Email: junzhe@quotr.ai

budgeting, and cost management. Price fluctuations, driven by raw material costs, macroeconomic conditions such as inflation and interest rates, and supply–demand imbalances, have amplified risks of cost overruns and delays (Abdul Nabi et al. 2024a; Crosthwaite 2000; Musarat et al. 2020; Shahhossein et al. 2017; Yang 2025).

Traditional econometric methods (i.e., multiple regression analysis) and modern econometric methods (i.e., univariate and multivariate time-series methods) have faced limitations in effectively capturing the high-frequency volatility observed in construction material prices (Shiha and El-adaway 2024). These models often struggle to handle the complexity of input data and exhibit limited predictive accuracy in real-world applications. As a result, advanced machine-learning methods have gained increasing attention. Among them, long short-term memory (LSTM) neural networks have demonstrated the capability to model complex temporal dependencies, offering improved accuracy compared with traditional methods. This motivates the development of a robust, granular, and scalable forecasting framework tailored to the construction domain.

## Problem and Challenge

To systematically address this problem, this study adopts the Construction Specifications Institute (CSI) MasterFormat as the organizing data structure. CSI MasterFormat provides a hierarchical classification system for construction materials and methods, with two-digit Divisions denoting broad categories and six-digit Sections specifying detailed materials and systems (ABC National, n.d.; Hall and Giglio 2013). Forecasting material prices at the six-digit Section level enables fine-grained cost estimation across a comprehensive range of building materials. In this study, the six-digit Section is treated as a standardized forecasting unit that provides a category-level input-cost signal: it is broad enough to average related items and reduce item-level noise, while retaining more material detail than two-digit Division-level indices.

However, developing such a framework presents two major challenges. First, assembling a reliable dataset requires integrating multiple material categories with at least a decade of temporal coverage to capture historical price dynamics. Second, constructing an effective predictive model demands careful selection of algorithms and feature engineering strategies capable of capturing nonlinear relationships among explanatory variables, such as commodity indexes and macroeconomic indicators. Overcoming these challenges is essential for improving budgeting reliability, procurement planning, and overall cost management in construction projects.

## Literature Review

As stated above, construction material cost forecasting has received growing attention. In this section, we offer an extensive analysis of existing research in the field of construction material cost forecasting: (1) material cost estimation; (2) econometric time-series approaches; and (3) machine-learning model frameworks.

Within standard cost estimate taxonomies, our study is positioned at the engineer's and definitive estimate levels rather than conceptual or order-of-magnitude stages (Jafari and Mousavi 2026). In this study, the definitive level refers to later-stage detailed estimating, where quantities and specification details are sufficiently developed for granular unit-cost inputs. Parametric and assembly system models support conceptual and intermediate estimates under limited design definition, whereas unit price and schedule estimates are typically prepared later in design development or during the construction documents phase, once detailed quantities, assemblies, technical specifications, and scope requirements are sufficiently defined. These latter estimates correspond to engineer's and definitive estimates and offer the highest accuracy but are also the most data-intensive, relying on granular bills of quantities and historically calibrated unit costs. Nguyen's process-based cost modeling (PBCM) reframes detailed estimating as rapid feedback for target value design, making both product and process costs (logistics, sequencing, site handling) explicit to reduce contingencies and support value-driven design iteration (Nguyen et al. 2018).

Empirical work in the Journal of Construction Engineering and Management similarly underscores how the quality of engineer's estimates and the treatment of input price uncertainty affect project outcomes. Gransberg and Riemer (2009) demonstrate that inaccuracies in engineer's estimated quantities materially affect unit price contracts, inducing unbalanced bids and cost growth when quantity risk is mischaracterized. Against this backdrop, our granular forecasting of residential construction material prices is intended to strengthen engineer's and definitive estimates by supplying statistically grounded, forward-looking unit prices for key commodities, thereby reducing reliance on ad hoc escalation allowances and improving the robustness of late design cost commitments (Gransberg and Riemer 2009).

Traditional construction cost estimation has often relied on transparent parametric or regression-based models at the project level. In this study, our focus is narrower: forecasting pre-markup construction input costs at the CSI six-digit Section level rather than final bid or total project costs. While regression-based approaches remain interpretable and data-efficient, project-level models are not designed to capture nonlinear material price dynamics and heterogeneous macroeconomic drivers at this level of granularity. This motivates the use of time-series and machine-learning forecasting approaches for Section-level material price prediction. At the material level, econometric analyses have investigated how commodity price shocks transmit into construction producer price indices, with macroeconomic variables such as consumer price index (CPI), gross domestic product (GDP), and employment serving as proxies for inflationary pressures (Chakraborty et al. 2020; Habib et al. 2025; Wang et al. 2022). While regression-based methods remain transparent and data-efficient, they struggle to capture nonlinear interactions, transfer poorly across building types and market conditions, and are generally confined to project- or index-level applications. Importantly, they lack the granularity required for material-level price prediction, leaving a gap for more advanced modeling approaches.

Econometric time-series methods are widely used to forecast construction costs and material prices. Univariate Box–Jenkins models, such as autoregressive integrated moving average (ARIMA) and seasonal autoregressive integrated moving average (SARIMA), provide classical time-series forecasting baselines. Automated ARIMA selection procedures make these models practical for repeated series-level applications (Hyndman and Khandakar 2008). However, standard univariate specifications do not directly incorporate exogenous macroeconomic or sectoral drivers unless extended, motivating the use of multivariate specifications when such indicators are informative. Recent ASCE research has extended multivariate forecasting to selected construction materials. Shiha and El-adaway (2024) forecasted construction material prices using macroeconomic indicators of trading partners, further underscoring the value of exogenous macroeconomic information in construction material price prediction (Shiha and El-adaway 2024). Vector autoregression (VAR) and vector error-correction models (VECM) capture both short-run adjustments and long-run cointegration among construction indices and their determinants, and have been validated on highway cost indices and individual material prices (Altalhoni et al. 2025; Shahandashti and Ashuri 2016). These econometric frameworks require careful lag-order selection and formal stationarity/cointegration testing (e.g., Johansen procedures) (Shahandashti and Ashuri 2016). Mixed-frequency approaches such as mixed data sampling (MIDAS) allow monthly producer price indices to be incorporated into quarterly construction forecasts without temporal aggregation, offering efficiency gains over standard distributed-lag models (Clements and Galvão 2008). Evaluation protocols typically involve rigorous out-of-sample tests, such as rolling-origin validation, combined with accuracy metrics for fair multihorizon comparisons at 1-, 2-, and 4-quarter-ahead targets. Even so, standard time-series specifications are often strained by nonlinear responses, structural breaks, and high-dimensional exogenous information. These limitations motivate the exploration of machine-learning frameworks.

Machine-learning approaches complement econometric models by capturing nonlinear interactions and high-dimensional feature spaces, with recent studies showing consistent improvements over linear baselines when models are properly tuned and validated (Chakraborty et al. 2020; Habib et al. 2025; Ottaviani et al. 2025). Tree-based ensembles such as random forests and gradient boosting are widely applied for mixed economic–technical features; their variance-reduction mechanisms (bagging and decorrelated splits) make them robust under noisy signals, and they have repeatedly outperformed regression-based methods in construction cost prediction (Breiman 2001; Habib et al. 2025; Ning et al. 2025; Shoar et al. 2022). Recent work extends this line to construction material price prediction, with Ning et al. (2025) proposing a sliding-window CatBoost model for this task (Ning et al. 2025). Sequence models directly encode temporal dependencies, and LSTM architectures and related hybrids such as exponential smoothing–recurrent neural network (ES-RNN) have achieved competitive accuracy across diverse series and horizons, making them suitable for material price forecasting where lagged effects and regime persistence are important (Cheng et al. 2025; Wang et al. 2026). At the broader cost-index level, Wang et al. (2026) also explored more advanced hybrid deep-learning architectures, such as the variational mode decomposition-gated recurrent unit (VMD-GRU) framework with the multihead self-attention (MHSA) mechanism. In addition, uncertainty quantification has become increasingly emphasized for budgeting and procurement, with interval and probabilistic approaches, such as ANN-based lower–upper bound estimation and quantile regression, providing decision-useful confidence bands beyond single-point predictions (Chen et al. 2025; Mir et al. 2021). Jezzini et al. (2025) further advanced this direction by linking predictive probabilistic deep-learning algorithms with stochastic risk modeling to quantify project cost risks due to construction material price volatilities. Reflecting these developments, material-level forecasting at CSI-like granularity typically engineers lagged and rolling statistics from producer price and macroeconomic indicators, applies both tree ensembles and recurrent networks, and evaluates models using multihorizon accuracy together with interval quality to balance predictive performance, transparency, and risk communication (Chakraborty et al. 2020; Chen et al. 2025; Habib et al. 2025; Mir et al. 2021; Wu et al. 2025). To further explore Transformer-based approaches for time-series forecasting, a vanilla Transformer architecture was also considered, following Caetano et al. (2025). The vanilla Transformer enables the evaluation of standard attention-based sequence modeling under limited data conditions while explicitly accounting for explanatory variables. It also provides a baseline for comparing Transformer-based performance against both econometric models and more advanced architectures. Lastly, we include Chronos-Bolt, the domain transfer time-series model introduced by Ansari et al. (2024) to exploit cross-sector signals for zero-shot prediction. This method's capabilities have not previously been published for material prices or construction costs; while performance is uncertain, we consider it a worthwhile exploratory addition (Hornek et al. 2025).

In summary, despite advances in regression-, econometric time-series-, and machine-learning approaches, several important gaps remain:

1. Gap 1 (Granularity): Most prior studies focus on project-level models or narrow sets of materials, lacking the granularity needed for material-level forecasting under standardized classification systems such as CSI.
2. Gap 2 (Methodological limitations): Traditional time-series models such as ARIMA and VECM rely on restrictive assumptions and struggle with nonlinear dependencies, structural breaks, and high-dimensional exogenous variables.
3. Gap 3 (Practical decision support): While machine-learning approaches have improved predictive accuracy, they are rarely applied at the CSI six-digit level and often neglect uncertainty quantification and interpretability, which are essential for decision-making in procurement and budgeting.

These limitations highlight the need for a granular, scalable, and decision-oriented forecasting framework, motivating the present study.

## Contributions

The present study develops a new forecasting framework that directly responds to the identified gaps. The contributions can be summarized as follows:

1. Granular dataset construction (addressing Gap 1): We assemble a CSI six-digit, multiseries corpus covering 22 Divisions with quarterly RSMeans material prices from Q12007–Q22025 (78 observations per series), enabling Section-level forecasting across multiple material categories at a level of granularity rarely examined in prior construction cost forecasting research.
2. Methodological validation (addressing Gap 2): We design and evaluate a consistent forecasting framework that compares representative approaches, ARIMA, VECM, LSTM, vanilla Transformer and Chronos-Bolt, under common training and evaluation protocols. This allows the study to identify their relative strengths and weaknesses in a short-history, heterogeneous construction-price setting and to show the value of incorporating producer price index (PPI) and macroeconomic explanatory variables.
3. Practical decision support for construction (addressing Gap 3): We demonstrate that CSI Section-level material price forecasts can directly support developers' and definitive estimates by improving unit-cost updates, informing procurement timing, identifying exposure to volatile trades, and strengthening budget reviews during preconstruction decision-making.

## Overview of the Proposed Framework

To provide a clear overview of the study design, Fig. 1 presents the overall workflow of the proposed construction material price forecasting framework. The framework consists of four major stages: data collection, preprocessing and feature engineering, model development under baseline and augmented settings, and model evaluation and selection. Detailed descriptions of each stage are provided in the following sections. The framework integrates multiple data sources, including RSMeans CSI data, PPI raw material data, macroeconomic indicators, and city- and commodity-level indexes. After data preprocessing and feature engineering, forecasting models are developed under two settings: a baseline method using historical material price data only, and an augmented method incorporating explanatory variables. Multiple models are trained and tuned through fivefold cross-validation, and their predictive performance is compared using evaluation metrics such as bidirectional mean absolute percentage error (biMAPE), median absolute deviation (MedAE), directional accuracy, and prediction bias. The final stage summarizes model selection and limitation analysis.

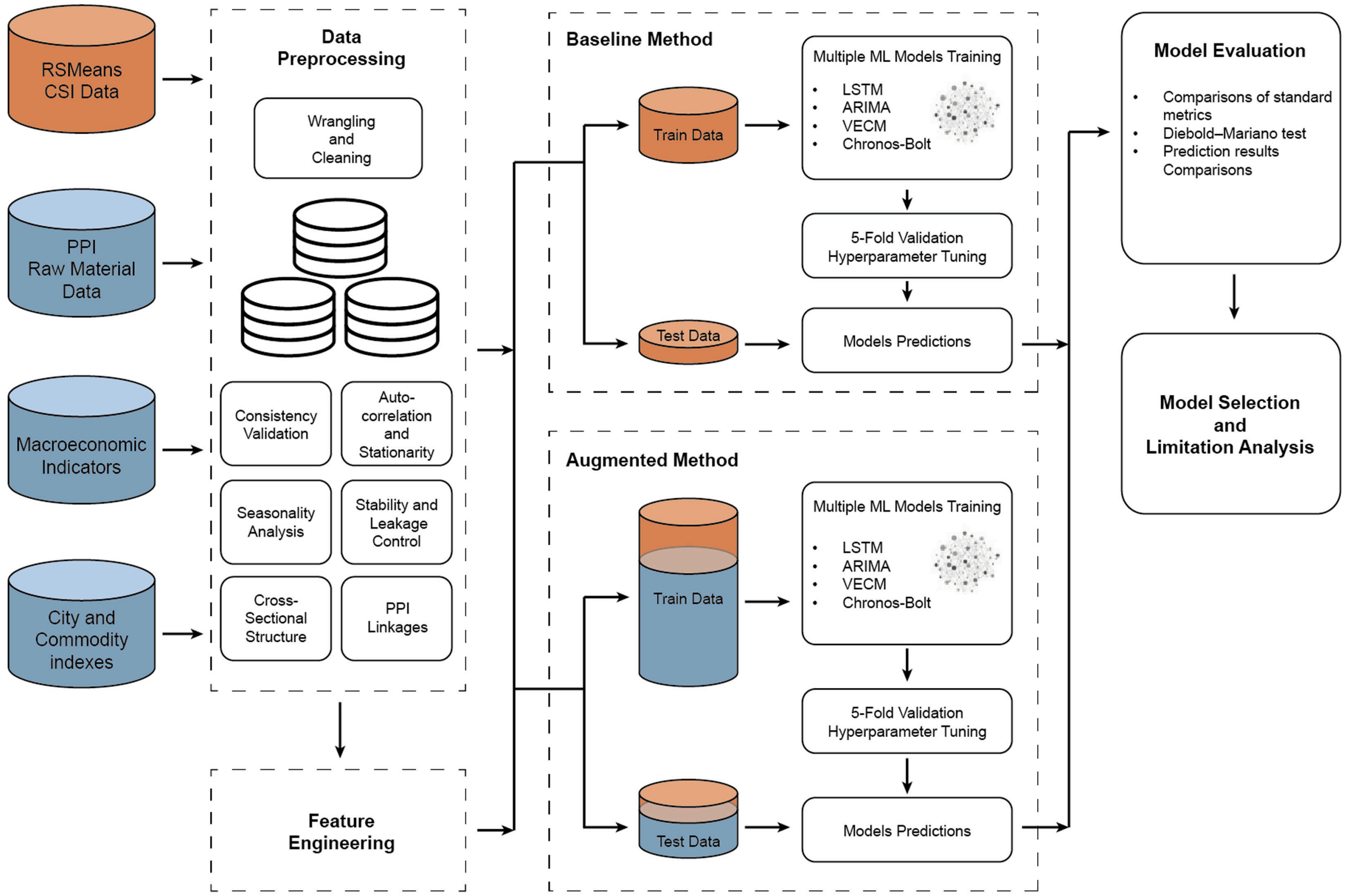

**Fig. 1.** Overall workflow of the proposed construction material price forecasting framework.

### *Organization of the Paper*

The remainder of this paper is organized as follows. Section "Data Description" introduces the dataset, including quarterly RSMeans material prices, producer price indices, and macroeconomic indicators, and explains how these data are processed for training, validation, and testing at the CSI six-digit level. Section "Methodology" presents the forecasting framework, detailing the implementation of ARIMA, VECM, LSTM, and Chronos-Bolt models with engineered lagged and rolling features. Section "Results and Discussion" reports and analyzes the experimental results, comparing baseline and extended model configurations. Section "Conclusion" concludes the paper by summarizing the key findings, discussing implications for construction cost management, and outlining directions for future research.

## Data Description

This section introduces the datasets used to develop and evaluate the proposed forecasting framework.

### *Data Source*

This study draws on three primary data sources to construct the forecasting framework: RSMeans Construction Cost Data, PPI series, and macroeconomic indicators, which are discussed separately as follows.

#### RSMeans Data

RSMeans Construction Cost Data is a widely recognized industry-standard resource for estimating construction costs in North America. Published annually by Gordian, it compiles detailed cost information for construction materials, labor, and equipment, organized according to the CSI MasterFormat. The MasterFormat classification system is hierarchical, with 2-digit Divisions representing broad categories of construction activities and 6-digit Sections specifying particular materials, methods, or systems (RSMeans 2012). In this context, Division 06 (Wood, Plastics, and Composites) is used as the primary demonstration division because of its central importance in residential construction and its strong relevance to framing, finishes, and other housing-related applications.

For this study, we construct a dataset from RSMeans annual publications spanning 2007–2025. The dataset contains 11,185 material-bearing items systematically mapped into 22 Divisions and 455 distinct 6-digit Sections. Each record includes material, labor, equipment, and installation costs, together with unit specifications and metadata, enabling analysis of both cross-sectional variation across materials and temporal evolution within Divisions. To ensure comparability with national cost levels, all values were normalized using the San Jose City Cost Index.

Figs. 2 and 3 illustrate the historical price index trends aggregated at the Division level and for Division 06 (Wood, Plastics,

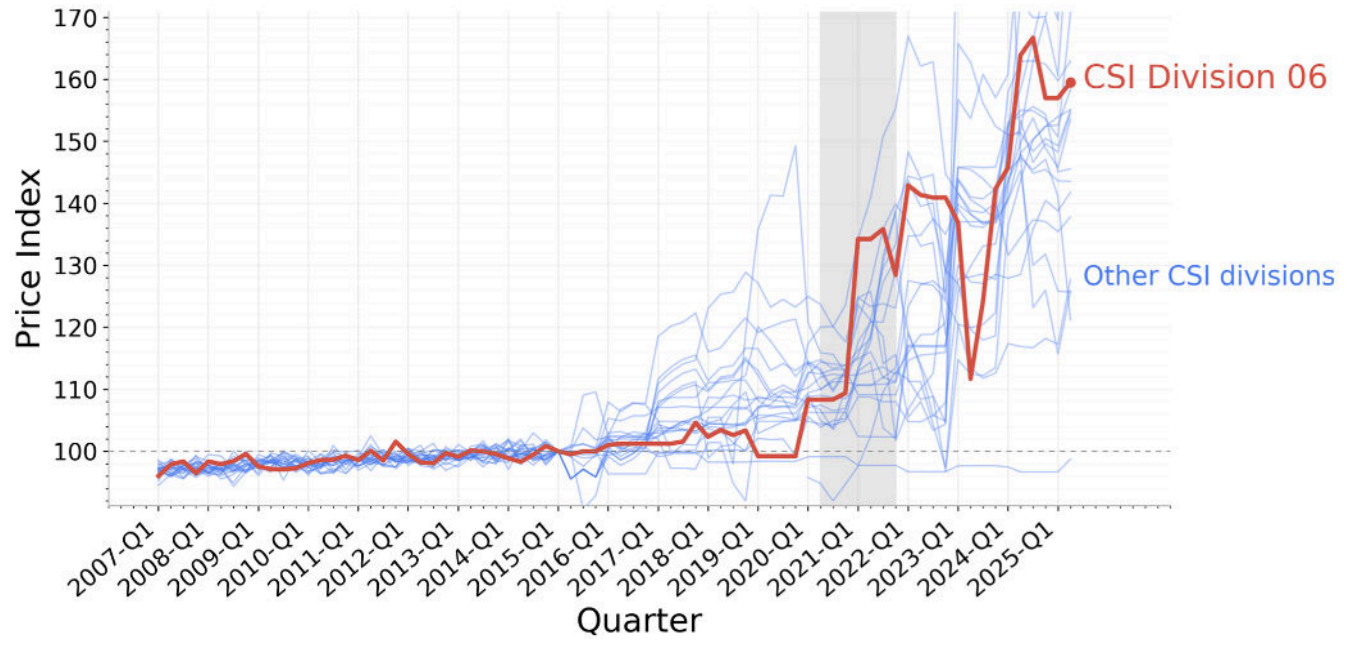

**Fig. 2.** Division-level price index trends, Q1 2007–Q1 2025.

and Composites), respectively. Division 06, which comprises 2,490 items across 35 subsections, is particularly significant in North American residential construction due to its role in structural framing, finishes, and exterior applications. Highlighting this Division provides a representative view of the broader dynamics captured in the RSMeans dataset, including material-specific volatility and long-term cost shifts relevant for construction planning and budgeting. The Fig. 2 shaded region denotes the COVID-19 shock window (Q4 2020–Q1 2022), when disruptions happened that drove a rapid change. The Fig. 2 shaded bands summarize cross-sectional dispersion across CSI six-digit Sections, the darker band shows the interquartile range around the median, while the lighter envelope traces the broader spread driven by sections with large price movements.

## Producer Price Index

To capture upstream commodity price movements, this study incorporates PPI data published by the US Bureau of Labor Statistics (BLS) (Bureau of Labor Statistics, n.d.-a). The PPI measures the average change over time in the selling prices received by domestic producers and is widely used as an indicator of inflation at the wholesale level. Construction-related PPIs cover key inputs such as lumber, steel, concrete, plastics, and glass (Akcay 2011), making them directly relevant for construction cost analysis.

The PPI program includes data that track most of the sectors of the economy, especially construction. This study utilizes a total of 145 distinct PPI series, each corresponding to a specific construction-related commodity. These include a wide range of materials and components commonly used in building projects, such as concrete, lumber, glass, metal, plastics, rubber, electronics, semiconductors, etc. The inclusion of this diverse set of indices allows for a nuanced examination of price movements across both traditional building materials and technologically advanced components.

Fig. 4 illustrates the aggregate trends of selected construction-related PPIs from 2007 to 2025, highlighting the heterogeneity of commodity price dynamics and their potential influence on construction material costs. These commodities encompass a broad range of materials and components commonly utilized in construction projects, including concrete, wood, glass, metals, plastics, and rubber.

## Macroeconomic Data

In addition to RSMeans Construction Cost Data (target variables) and PPI series (commodity-level features), this study incorporates a set of macroeconomic indicators to capture broader economic conditions influencing construction material costs. These variables reflect both localized and national dynamics that shape demand, inflation, and input costs. In this study, input cost refers to the direct cost of construction inputs recorded in RSMeans, excluding contractor overhead, profit, contingency, and other bid-related adjustments. At the local level, the city cost index (CCI) is included to account for regional variations in labor, material, and equipment costs relative to a national benchmark.

At the national and sectoral levels, we incorporate a range of macroeconomic indicators drawn from the US BLS (Bureau of Labor Statistics, n.d.-b), Bureau of Economic Analysis (BEA) (Bureau of Economic Analysis, n.d.), and US Census Bureau. These include:

- Inflation and labor market conditions: CPI, employment in construction (EC), employment rate (ER), and hourly wages (HE).
- Economic activity and purchasing power: GDP, personal income (PI), and total construction spending (CS).
- Construction demand: housing starts (HS) and building permits (BP).
- Commodity and energy markets: global iron ore price (IOP) and West Texas intermediate (WTI) crude oil.

Together, these indicators provide explanatory context for both cyclical and structural drivers of material price changes, complementing commodity-specific PPIs. Fig. 5 illustrates the historical trajectories of these macroeconomic indicators from 2007 to 2025, highlighting the broader economic trends that shape construction cost dynamics.

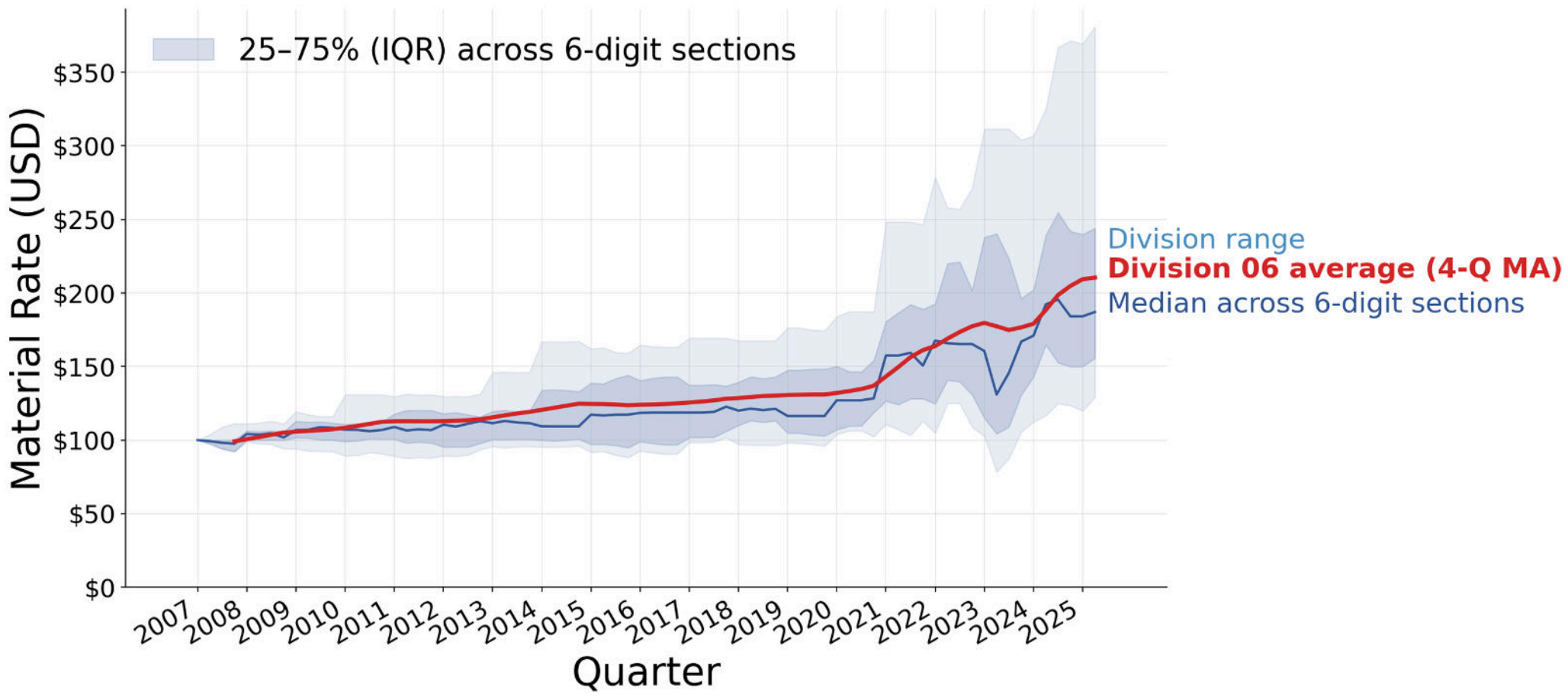

**Fig. 3.** Division 06 Wood, Plastics, and Composites subsections, 2007–2025.

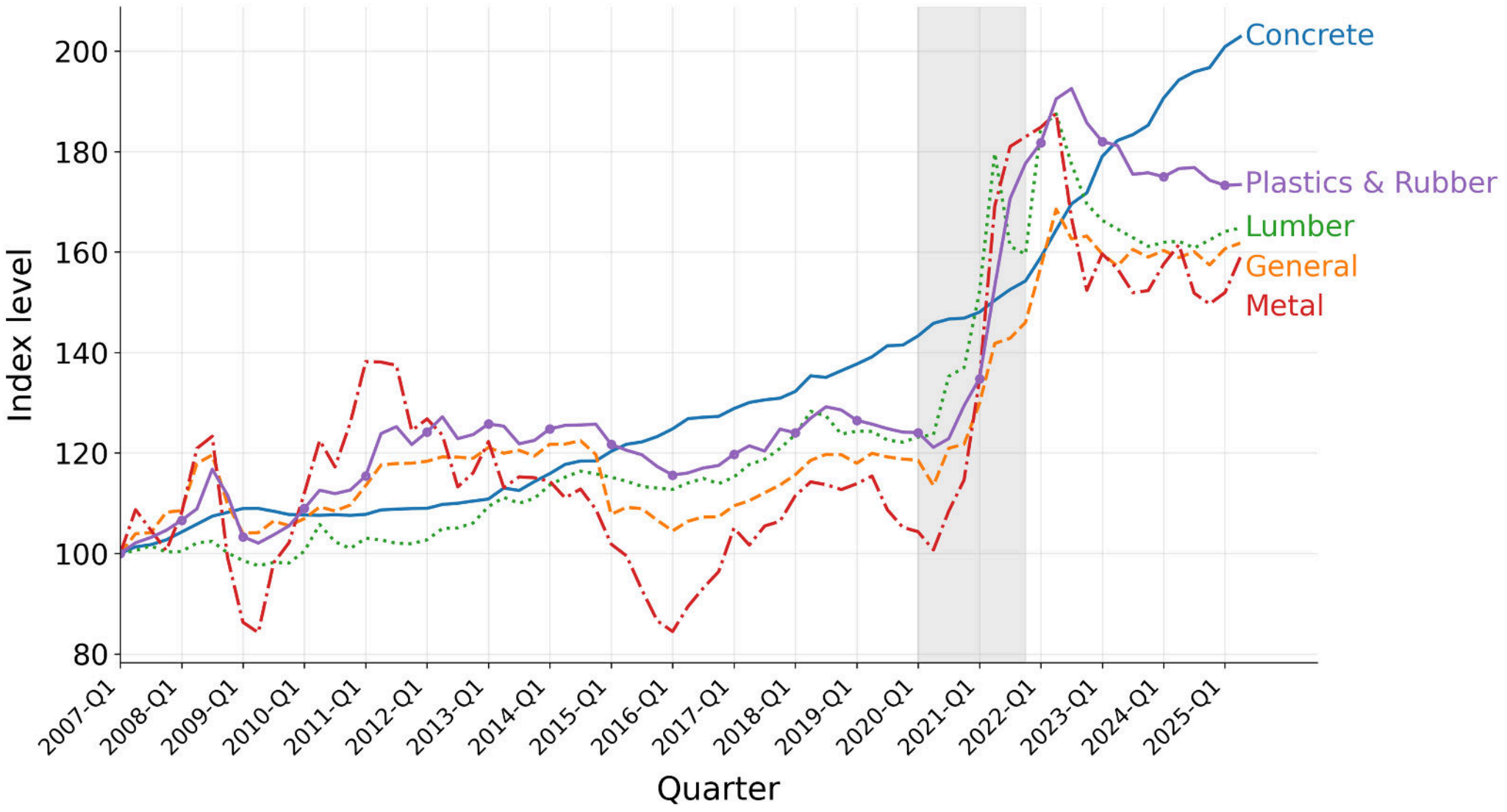

**Fig. 4.** Price index trends for selected construction-related PPI commodities, 2007–2025.

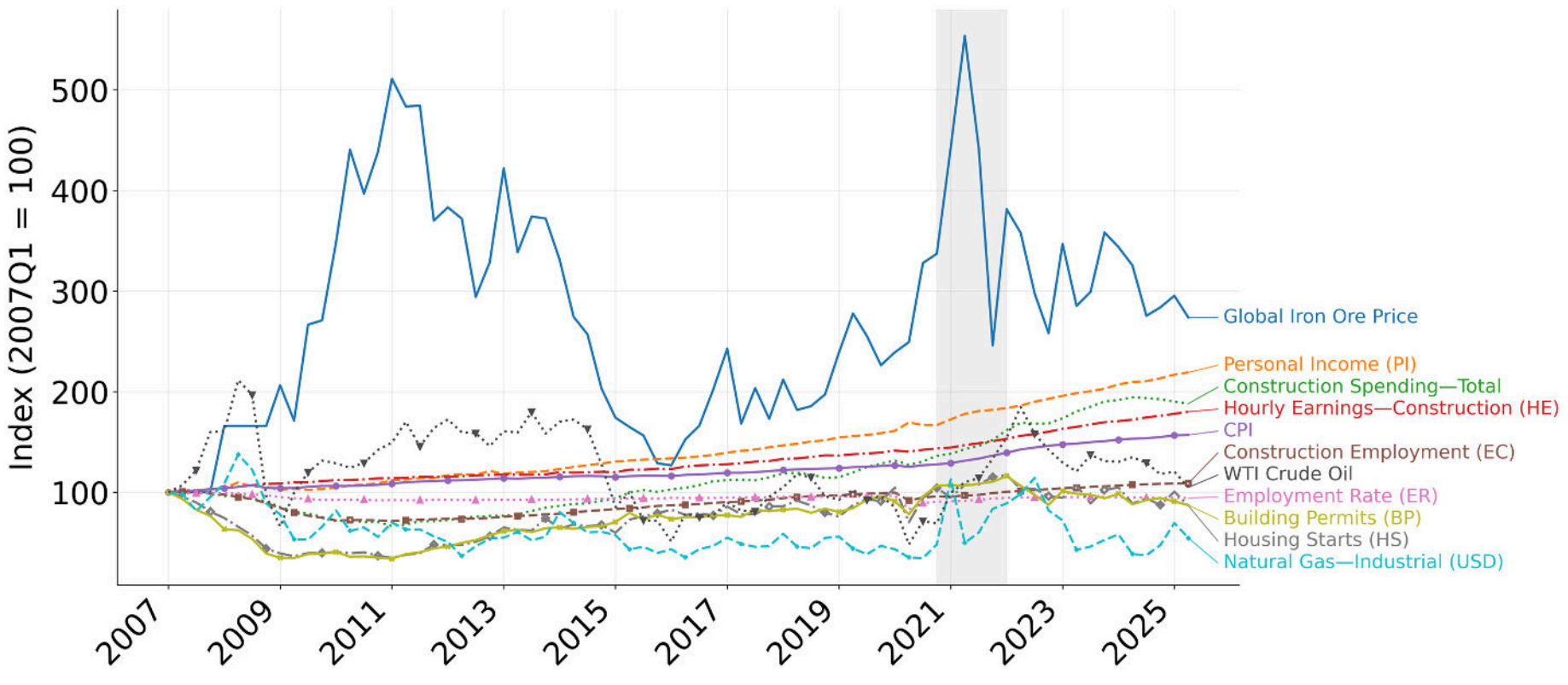

**Fig. 5.** Historical trends of selected macroeconomic indicators, 2007–2025.

## *Data Preprocessing*

Before applying forecasting models, all datasets were preprocessed to ensure consistency, comparability, and predictive relevance. This subsection describes the key steps, including aggregation and normalization, exclusion of non-material content, and the train–test split design. In addition, exploratory data analysis (EDA) was conducted to validate preprocessing and to characterize important features of the dataset, such as seasonality, autocorrelation, and correlations with explanatory variables.

### Wrangling and Cleaning

All data were aggregated and processed on a quarterly basis to capture seasonal patterns and short-term market fluctuations in construction material prices (Choi et al. 2021). To facilitate temporal comparison and remove the influence of absolute price levels, all material price data were standardized by setting the first quarter of 2007 as the baseline, with an index value of 100. This normalization enables consistent evaluation of relative price changes over time and allows for direct comparison across different material categories and cost divisions.

We restrict the corpus to material-bearing Sections and exclude service-type content. Specifically, we remove Division 01, General Requirements, Division 02, and Existing Conditions from modeling as these predominantly capture administrative, project management, and site-service activities that lack direct material price mapping to PPIs. Across other Divisions, we drop rows flagged as services and overheads using RSMeans metadata, including Division and Section titles and descriptions, and a concise keyword taxonomy (e.g., mobilization, project management, temporary facilities, permits/fees, bonding, testing/inspection, documentation, submittals, etc.). This exclusion prevents conflating negotiated or project-specific charges with commodity-linked price signals, preserving comparability to PPIs, economic indicators, and CCI adjustment, avoiding spurious lead-lag inferences. Suspicious extreme observations in the raw RSMeans-based data were manually reviewed case by case; obvious source-data errors were corrected, whereas economically plausible extreme movements were retained.

### Exploratory Data Analysis

EDA was conducted on the preprocessed dataset for material price from 2007 to 2025 to validate data integrity and to characterize temporal and cross-sectional features relevant for forecasting. The analysis focused on six aspects: consistency validation, seasonality, autocorrelation and stationarity, cross-sectional structure and PPI linkages, stability, and feature correlations.

*Consistency validation*: Table 1 reports cross-sectional dispersion before and after applying the CCI. Aggregating RSMeans

**Table 1.** Cross-sectional dispersion of CSI six-digit Sections after CCI adjustment

| Quarter | Number of sections | Non-null % | Median | IQR | Mean | SD | CV |
|---|---|---|---|---|---|---|---|
| 2016_Q1 | 414 | 100.00 | 101.768 | 6.353 | 103.713 | 17.059 | 0.164 |
| 2019_Q1 | 412 | 99.52 | 108.233 | 15.261 | 111.455 | 20.950 | 0.188 |
| 2022_Q1 | 414 | 100.00 | 123.753 | 43.908 | 134.374 | 62.568 | 0.466 |
| 2025_Q1 | 414 | 100.00 | 157.346 | 78.163 | 236.379 | 424.311 | 1.795 |

Note: Post-CCI RSMeans indices aggregated to OSI 6-digit Sections. Representative recent quarters shown. Division 06 is rebased to 201501 = 100. The EDA sample. CV = SD/Mean. IQR = interquartile range; SD = standard deviation; and CV = coefficient of variation.

data from item-level to Section-level (6-digit) reduced noise while preserving quarter-to-quarter dynamics, supporting the decision to model at the Section level.

*Seasonality*: Seasonal decomposition [seasonal and trend decomposition using Loess (STL) and dummy regressions] revealed systematic intrayear fluctuations that were material-specific rather than uniform across Divisions. Wood-related Sections showed Q1 peaks, consistent with framing and procurement cycles, while metals and plastics exhibited later peaks. Autocorrelation diagnostics confirmed seasonal lags at 4, 8, and 12 quarters (Fig. 6).

*Cross-sectional structure and PPI linkages*: Correlation clustering showed that Sections were grouped into intuitive input markets (e.g., wood, metals, plastics). Cluster prototypes summarized shared dynamics and revealed contagion during volatility episodes (Abdul Nabi et al. 2024b). Cross-correlations with PPI series indicated strong contemporaneous associations, often with PPIs leading at short horizons, motivating the inclusion of lagged PPI features (Fig. 7).

*Stability and leakage control*: Rolling-window correlations demonstrated that Section-PPI relationships were generally persistent but varied across regimes. This justified the use of rolling-origin evaluation and regularization. To ensure strict temporal integrity and prevent data leakage, the proposed framework adopts a fully causal time-series modeling pipeline. Specifically, the dataset is partitioned using a chronological split, where training data strictly precedes validation and test data, without any random or stratified sampling across the temporal range. All preprocessing steps, including scaling, are fitted exclusively on the training set and then applied to subsequent data. Feature engineering techniques (e.g., lag features, rolling statistics, and differencing) are constructed using only historical observations, ensuring that no future information is introduced.

During model training and evaluation, future data are never exposed to the model. Predictions are generated prior to observing the corresponding ground-truth values, and model updates are performed only after new observations become available. This walk-forward procedure follows a realistic deployment scenario (predict → observe → update → predict next), guaranteeing that the model operates strictly within the information available at each time step. All features were constructed using only historical information available at each timestamp, ensuring leakage-free modeling (Fig. 8).

*Feature correlations*: Pearson, Spearman, and sum-of-squared-residuals (SSR) metrics were used to screen explanatory features. Features with narrow distributions centered on high correlations were identified as strong predictors, while broader distributions around low values indicated weaker predictive relevance (Fig. 9).

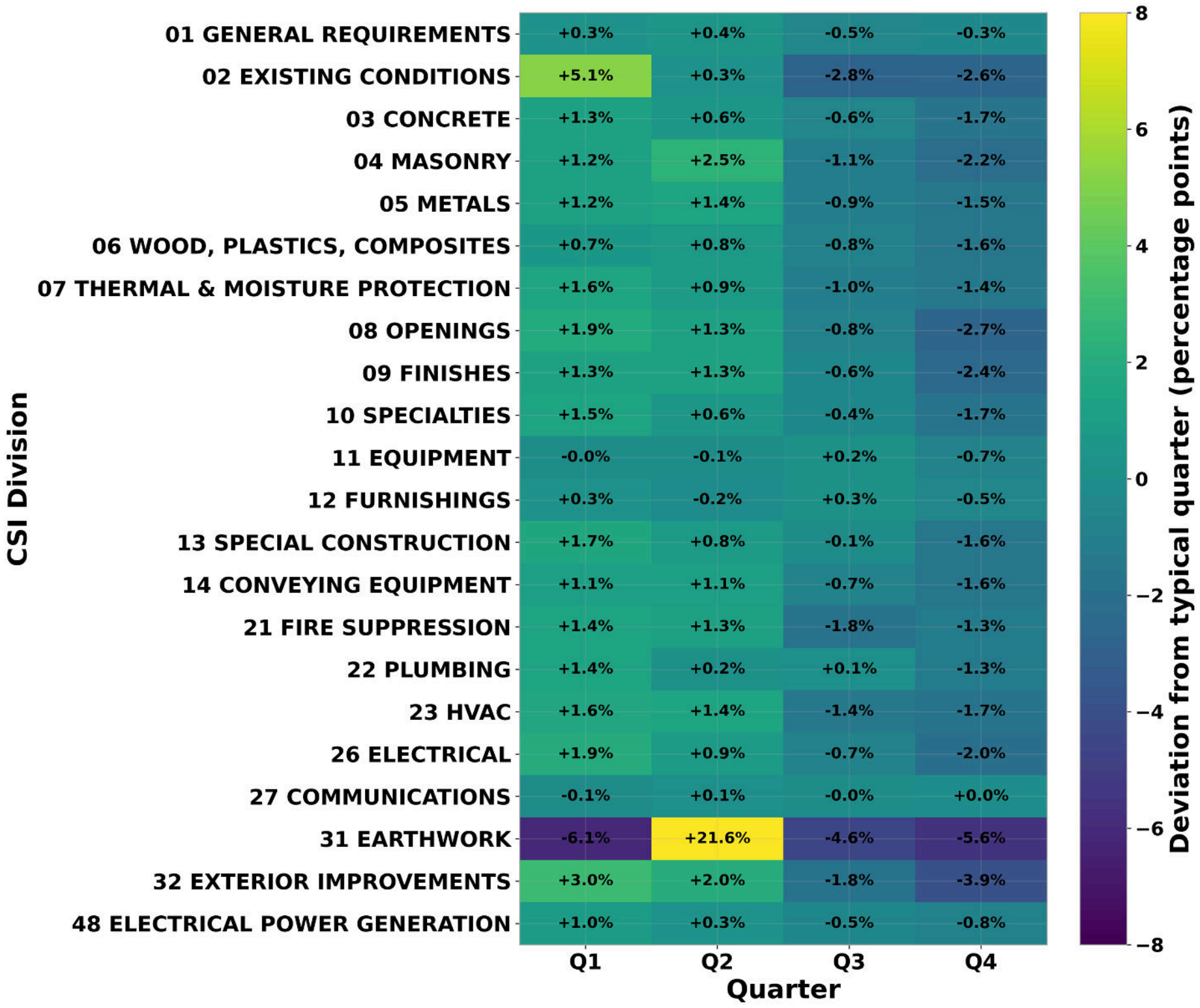

**Fig. 6.** Quarterly seasonality indices (Division-level seasonal indices (index-1), annotatedEDA window: Q1 2016–present. Div 06 base Q1 2015 = 100).

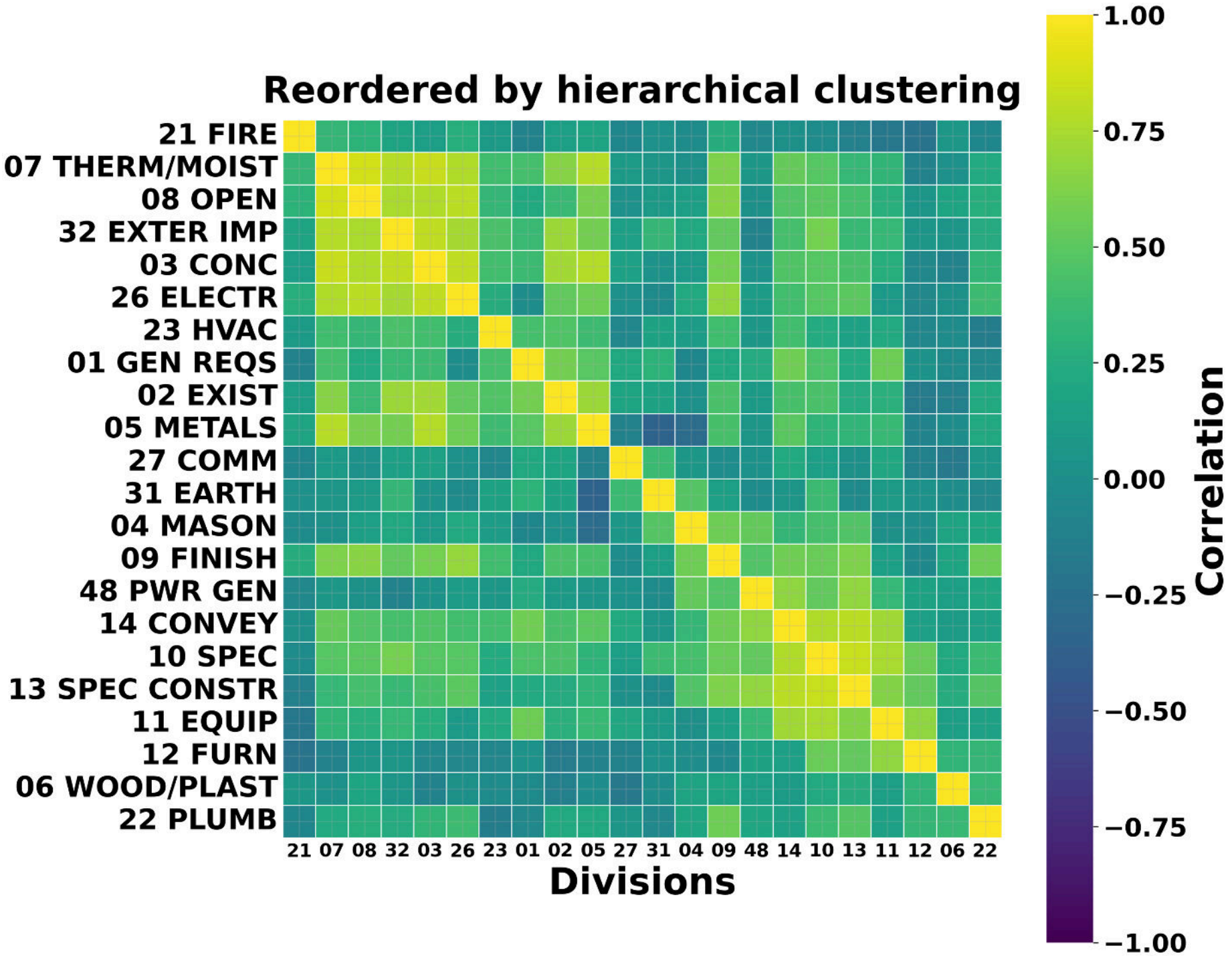

**Fig. 7.** Cross-sectional correlation structure among CSI Divisions (quarter-over-quarter log price changes).

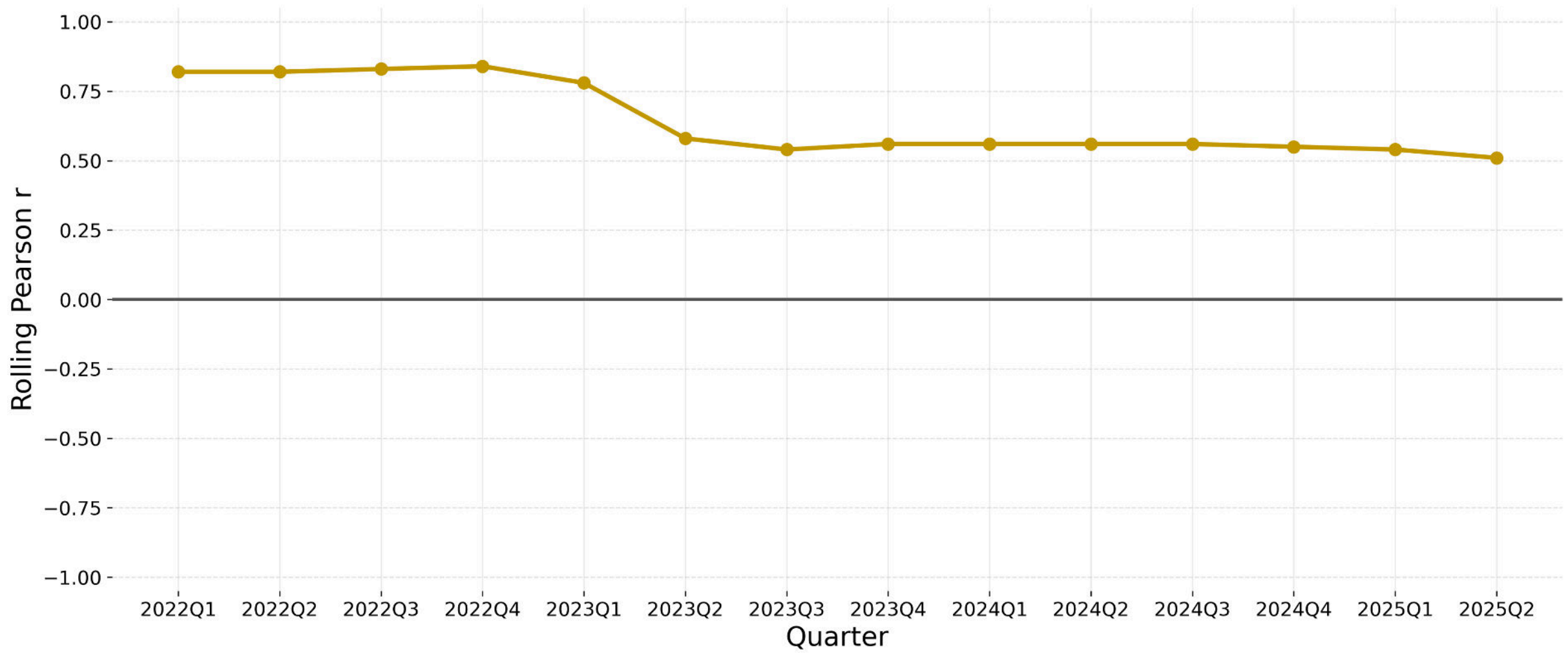

**Fig. 8.** Rolling 24-quarter correlations between Section-level and Division prototype indices.

In summary, the EDA confirmed that the dataset exhibits strong seasonal patterns, persistent autocorrelation, meaningful Section-PPI linkages, and heterogeneous explanatory power across features. These findings highlight the need for advanced forecasting models capable of capturing both temporal dynamics and cross-sectional dependencies, motivating the methodology presented in the section "Methodology."

## Methodology

This section outlines the forecasting framework developed in this study, encompassing model specification, training procedures, cross-validation design, and hyperparameter tuning, with both section-specific and unified models considered.

### Model Methodology

This study aims to forecast construction material price movements at the CSI six-digit Section level using a set of local (section-wise) multivariate time-series models, where each model forecasts one section index using its own history and a common pool of exogenous predictors. We consider quarterly RSMeans price indices for $N$ CSI six-digit Sections observed over $T$ quarters. For section $i \in \{1, \ldots, N\}$ and quarter $t \in \{1, \ldots, T\}$, let $y_{i,t} > 0$ denote the rebased material price index (e.g., with 2007Q1 as base period). Collecting all sections at time $t$ gives the $N$-dimensional vector

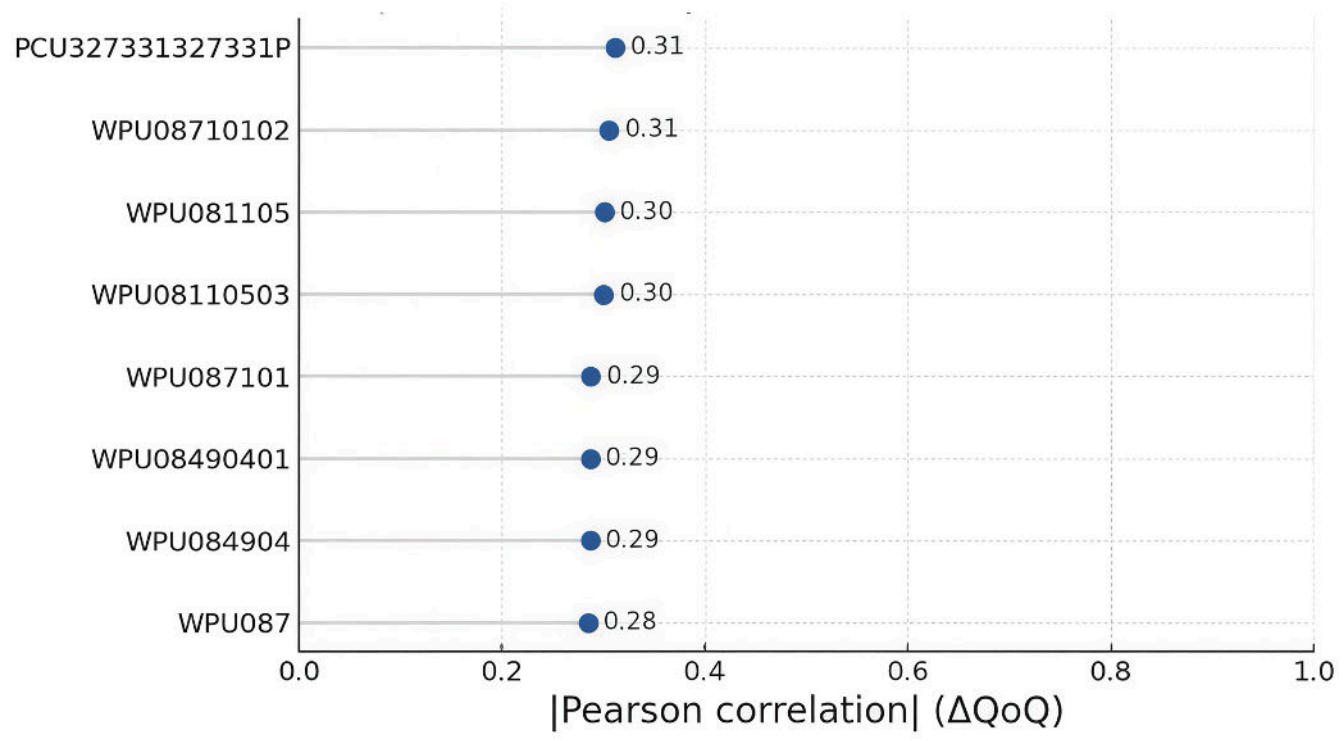

**Fig. 9.** Correlation of CSI Section price indices with top explanatory features (PPIs and macroeconomic indicators).

$$y_t = (y_{1,t}, \ldots, y_{N,t})^T \in R^N \tag{1}$$

where $y_t$ = vector of all CSI six-digit Section price indices at quarter $t$; $y_{i,t}$ = rebased material price index for CSI six-digit Section $i$; and $N$ = number of CSI six-digit Sections

$$\hat{y}_{i,t+h|t} = f(\{y_{i,t-\tau}, x_{j,t-\tau}: \tau = 0, \ldots, L-1; \,|\, j = 1, \ldots, J\}) \tag{2}$$

where $\hat{y}_{i,t+h|t}$ = forecast for CSI six-digit Section $i$ at quarter $t+h$, made at quarter $t$; $f(\cdot)$ = forecasting function; $L$ = lookback length; $x_{j,t-\tau}$ = explanatory variable $j$ observed $\tau$ quarters before $t$; and $J$ = number of explanatory variables

$$X_t = (x_{1,t}, \ldots, x_{J,t})^\top, \quad x_{j,t} \in \{x_{j,t}^{\mathrm{PPI}}, x_{j,t}^{\mathrm{macro}}, x_{j,t}^{\mathrm{CCI}}\} \tag{3}$$

where $X_t$ = vector of all explanatory variables available at quarter $t$; $x_{j,t}^{\mathrm{PPI}}$ = PPI-based explanatory variable; $x_{j,t}^{\mathrm{macro}}$ = macroeconomic explanatory variable; and $x_{j,t}^{\mathrm{CCI}}$ = city-level cost-index explanatory variable

$$\mathcal{I}_{i,t}^{\mathrm{base}}(L) = \{y_{i,t-\tau}: \tau = 0, \ldots, L-1\} \tag{4}$$

where $\mathcal{I}_{i,t}^{\mathrm{base}}(L)$ = base information set for CSI six-digit Section $i$, containing only historical target values; and $\tau$ = lag index

$$\hat{y}_{i,t+h|t}^{(m)} = f_h^{(m)}(\mathcal{I}_{i,t}(L); \theta_i^{(m)}), \quad i = 1, \ldots, N \tag{5}$$

where $\hat{y}_{i,t+h|t}^{(m)}$ = forecast from model $m$; $f_h^{(m)}(\cdot)$ = horizon-specific forecasting function; $\mathcal{I}_{i,t}(L)$ is either the base or augmented information set; and $\theta_i^{(m)}$ = parameters estimated for CSI six-digit Section $i$

$$y_{i,t+h} = \hat{y}_{i,t+h|t}^{(m)} + \varepsilon_{i,t+h}^{(m)} \tag{6}$$

where $y_{i,t+h}$ = realized material price index for CSI six-digit Section $i$; $\hat{y}_{i,t+h|t}^{(m)}$ = forecast; and $\varepsilon_{i,t+h}^{(m)}$ = zero-mean forecast error

$$\mathcal{I}_{i,t}^{\mathrm{aug}}(L) = \{y_{i,t-\tau}, x_{j,t-\tau}: \tau = 0, \ldots, L-1; \,|\, j = 1, \ldots, J\} \tag{7}$$

where $\mathcal{I}_{i,t}^{\mathrm{aug}}(L)$ = augmented information set for CSI six-digit Section $i$, containing historical target values and explanatory variables; and $j$ indexes PPI, macroeconomic, and CCI variables

$$y_{i,t} > 0, \quad x_{j,t}^{\mathrm{PPI}} > 0, \quad \mathbb{E}(y_{i,t}^2) < \infty, \quad \mathbb{E}(x_{j,t}^2) < \infty \tag{8}$$

where $y_{i,t}$ = CSI six-digit Section-level material price index; $x_{j,t}^{\mathrm{PPI}}$ = PPI-based explanatory variable; positivity ensures logarithmic transformations are well defined; and finite second moments support mean square error–based evaluation.

### Feature Selections

We construct features to reflect material-level dynamics, input-market signals, and macroeconomic context, using a fixed three-year (12-quarter) lookback horizon for all forecasts. RSMeans six-digit CSI Section indices serve as both the foundation for feature engineering and the target variables. Let

$$\tilde{y}_{i,t} = g(y_{i,t}) \tag{9}$$

where $\tilde{y}_{i,t}$ = transformed target value for CSI six-digit Section $i$; and $g(\cdot)$ = transformation function

$$\tilde{y}_{i,t}^{\log} = \log(y_{i,t}) \tag{10}$$

where $\tilde{y}_{i,t}^{\log}$ = log-transformed target for CSI six-digit Section $i$; $\log(\cdot)$ = natural logarithm; and $y_{i,t} > 0$ is required

$$D_{q,t} = \mathbb{1}\{Q(t) = q\}, \quad q = 1, 2, 3, 4 \tag{11}$$

where $D_{q,t}$ = seasonal dummy variable; $Q(t)$ returns the calendar quarter of time $t$; and $\mathbb{1}\{\cdot\}$ = indicator function equal to 1 when the condition is true and 0 otherwise

$$y_{i,t-1}, |y_{i,t-2}, |\ldots, |y_{i,t-12} \tag{12}$$

where these are lagged target values for CSI six-digit Section $i$, from one quarter to twelve quarters before quarter $t$

$$\Delta_4 y_{i,t} = y_{i,t} - y_{i,t-4} \tag{13}$$

where $\Delta_4 y_{i,t}$ = four-quarter, or year-over-year, change in the material price index for CSI six-digit Section $i$

$$\mathrm{MA}_{i,t}^{(k)} = \frac{1}{k} \sum_{\tau=0}^{k-1} y_{i,t-\tau}, \qquad \mathrm{SD}_{i,t}^{(k)} = \sqrt{\frac{1}{k-1} \sum_{\tau=0}^{k-1} (y_{i,t-\tau} - \mathrm{MA}_{i,t}^{(k)})^2} \tag{14}$$

where $\mathrm{MA}_{i,t}^{(k)}$ = moving average for CSI six-digit Section $i$ over a $k$-quarter window; $\mathrm{SD}_{i,t}^{(k)}$ = rolling standard deviation; $k$ = rolling-window length; and $\tau$ i = lag index.

To capture input-market dynamics, we use construction-relevant PPI series $x_{j,t}^{\mathrm{PPI}}$ (wood, metals, plastics, cement, etc.). Monthly PPIs are aligned to quarters, rebased, and transformed into levels, and lags up to 12 quarters, with cluster aggregates (e.g., "wood," "metals") smoothing idiosyncratic noise. Macroeconomic conditions enter through $x_{j,t}^{\mathrm{macro}}$ (CPI, GDP, income, construction employment, housing activity, construction spending, wages, and energy/commodity benchmarks), processed with the same transformations. External PPI and macroeconomic indicators were harmonized to the quarterly CSI Section panel before model training, and series entering the augmented feature matrix were retained based on complete aligned histories. Model-side transformations and preprocessing, including lag, change, rolling, and seasonal feature generation, scaling, screening, and collinearity controls, were applied within the chronological training or expanding-window workflow. The external indicators are used as quarterly aligned historical covariates; strict release-date matching and real-time vintage reconstruction are treated as deployment refinements rather than as the focus of the present model-comparison experiment.

### Modeling and Training Framework

This section presents the modeling and training framework used to evaluate forecasting performance. All models are trained under a consistent protocol to ensure comparability. Two model specifications are considered: a base model, which uses only RSMeans data with engineered time-series features, and a final model, which incorporates additional exogenous predictors such as PPIs and macroeconomic indicators. Forecasts are generated at the CSI six-digit Section level using Section-specific models, as this approach consistently outperforms a single unified model across divisions. A fixed train–test split is adopted, with rolling cross-validation applied within the training data, and hyperparameter tuning performed to balance model complexity and generalization.

To represent a broad methodological spectrum, five models are selected for evaluation: (1) the LSTM network, (2) the ARIMA model, (3) the VECM, (4) the vanilla Transformer, and (5) Chronos-Bolt, a state-of-the-art Transformer-based forecasting framework. These methods span both classical econometric techniques and modern machine-learning architectures, enabling a comprehensive assessment of their relative strengths and limitations. Training all five models on the same dataset under uniform conditions allows us to benchmark predictive accuracy and robustness, thereby identifying the most effective approaches for construction material price forecasting. This comparative design provides insights not only into model performance but also into the trade-offs between interpretability, computational efficiency, and adaptability in real-world cost estimation settings.

#### LSTM

LSTM networks are a class of recurrent neural networks designed to overcome the vanishing gradient problem and capture long-term dependencies in sequential data (Abbasimehr and Paki 2022). Through gated memory units, LSTMs selectively retain, update, and output information across time steps, enabling them to model both short- and long-range temporal dynamics (Yu et al. 2019). This makes them well suited for forecasting tasks where historical patterns strongly influence future outcomes (Staudemeyer and Morris 2019)

$$s_{i,t} = [y_{i,t}, x_{1,t}, \ldots, x_{J,t}]^{\top}, \quad X_{i,t}^{\text{seq}} = (s_{i,t-T_{\text{lookback}}+1}, \ldots, s_{i,t}) \tag{15}$$

where $s_{i,t}$ = combined target-and-feature input vector for CSI six-digit Section $i$ at quarter $t$; $X_{i,t}^{\text{seq}}$ = LSTM input sequence; and $T_{\text{lookback}}$ = sequence length

$$\begin{aligned} f_r &= \sigma(W_f[h_{r-1}, s_{i,r}] + b_f) \\ \iota_r &= \sigma(W_\iota[h_{r-1}, s_{i,r}] + b_\iota) \\ o_r &= \sigma(W_o[h_{r-1}, s_{i,r}] + b_o) \\ \tilde{c}_r &= \tanh(W_c[h_{r-1}, s_{i,r}] + b_c) \\ c_r &= f_r \odot c_{r-1} + \iota_r \odot \tilde{c}_r \\ h_r &= o_r \odot \tanh(c_r) \end{aligned} \tag{16}$$

where $r$ = local recurrent time step; $f_r$, $\iota_r$, and $o_r$ = forget, input, and output gates; $c_r$ = cell state; $h_r$ = hidden state; $\sigma(\cdot)$ = sigmoid function; $\odot$ denotes element-wise multiplication; $W_f$, $W_\iota$, $W_o$, and $W_c$ are learnable weight matrices, and $b_f$, $b_\iota$, $b_o$, and $b_c$ are the corresponding biases. The input gate is written as $\iota_r$, not $i_r$, to avoid conflict with CSI six-digit Section index $i$

$$\hat{y}_{i,t+1|t}^{\text{LSTM}} = W_{\text{out}}\text{ReLU}(W_d h_{T_{\text{lookback}}} + b_d) + b_{\text{out}} \tag{17}$$

where $\hat{y}_{i,t+1|t}^{\text{LSTM}}$ = one-step-ahead LSTM forecast for CSI six-digit Section $i$; $h_{T_{\text{lookback}}}$ = final hidden state; $W_d$ and $b_d$ = dense layer; $W_{\text{out}}$ and $b_{\text{out}}$ = final regression output; and ReLU$(\cdot)$ = rectified linear activation.

The architecture consisted of a single LSTM layer with 512 hidden units (approximately 1.46 million trainable parameters), followed by dropout regularization (0.02), a dense layer with ReLU activation, and a final regression output. Models were trained for 100 epochs using the Adam optimizer with weight decay (Sagheer and Kotb 2019), corresponding to approximately 300 initial update steps. After constructing lags of 8 quarters, the effective sample size for training is 52 observations per division, with 10 walk-forward test windows. A walk-forward evaluation strategy was adopted, in which each forecast was appended to the sequence for subsequent predictions. To enhance adaptability, the model was incrementally fine-tuned for 50 epochs after each new observation (Nelson et al. 2017), adding approximately 150 steps per window (1,800 total update steps). To prevent overfitting, early stopping (patience = 5, min_delta = 0.001) was applied during initial training but was not triggered. For each random seed, we computed the median root mean squared error (RMSE) across all 32 Sections in Division 06. Across eight seeds, the mean of these per-seed median RMSE values was 4.21 (standard deviation: 0.26), indicating robust performance and low sensitivity to initialization. A representative run with RMSE of 4.19 was selected for reporting detailed results in the main text. This design reflects a realistic forecasting pipeline in which models are continuously updated as new information becomes available, improving robustness in dynamic environments.

We also conducted diagnostic analyses on the selected LSTM configuration using both training–validation loss curves and learning curves. The results show that the validation loss closely follows the training loss without significant divergence throughout training, indicating stable generalization. In addition, as the training set size increases, both training and validation errors decrease and converge without a widening gap, further confirming the absence of high variance. These observations demonstrate that, despite the relatively large model capacity, the chosen hyperparameter configuration remains appropriate for the dataset. The corresponding diagnostic plots are provided in Appendix I.

#### ARIMA

The ARIMA model is a classical statistical approach for time-series forecasting that combines autoregression, differencing, and moving-average terms to capture temporal dependencies (Shumway and Stoffer 2017). By balancing these components, ARIMA can model both short-term correlations and long-term trends in nonstationary data, providing an interpretable benchmark against which more complex machine-learning methods can be compared (Kontopoulou et al. 2023; Siami-Namini et al. 2018; Sirisha et al. 2022)

$$\begin{aligned} &\Phi_P(B^s)\phi_p(B)(1-B)^d(1-B^s)^D(y_{i,t} - \delta^{\top}X_t) \\ &\quad = \Theta_Q(B^s)\theta_q(B)\varepsilon_{i,t} \end{aligned} \tag{18}$$

where $B$ = backshift operator, so $By_{i,t} = y_{i,t-1}$; $p, d, q$ are nonseasonal ARIMA orders; $P, D, Q$ = seasonal ARIMA orders; $s$ = seasonal period, here $s = 4$ for quarterly data; $\phi_p(B)$ and $\theta_q(B)$ = nonseasonal autoregressive (AR) and moving average (MA) polynomials; $\Phi_P(B^s)$ and $\Theta_Q(B^s)$ = seasonal AR and MA polynomials; $\delta$ = coefficient vector for exogenous variables; and $X_t$ = explanatory variable vector.

In this study, we employ a seasonal autoregressive integrated moving average with exogenous regressors (ARIMAX) specification to incorporate both autoregressive dynamics and exogenous predictors. The exogenous variable matrix included engineered time-based covariates together with external explanatory factors such as PPIs and macroeconomic indicators, enabling the model to account for both intrinsic price dynamics and contextual drivers. To better approximate real forecasting conditions, a rolling-window strategy was used. At each step in the test horizon, the most recent 32 observations and their associated exogenous features were used to fit a seasonal ARIMAX (1,1,1) (1,1,1,4) model, which explicitly captured quarterly seasonal patterns. A one-step-ahead forecast was generated, and the window was then advanced by one period to incorporate the new observation into the training history. Formally, the one-step-ahead forecast at time $t + 1$ is:

$$\hat{y}^{\mathrm{ARIMAX}}_{i,t+1|t} = \mathbb{E}[y_{i,t+1} \mid y_{i,t-31:t}, \{x_{j,t-31:t}\}^{J}_{j=1}] \tag{19}$$

where $\hat{y}^{\mathrm{ARIMAX}}_{i,t+1|t}$ = ARIMAX forecast for CSI six-digit Section $i$; $y_{i,t-31:t}$ = most recent 32 target observations; $x_{j,t-31:t}$ = corresponding 32-quarter history of explanatory variable $j$; and $\mathbb{E}[\,\cdot \mid \cdot\,]$ = conditional expectation.

This walk-forward procedure ensured that forecasts remained responsive to local dynamics. To safeguard against convergence failures, a fallback mechanism was implemented. If the seasonal autoregressive integrated moving average with exogenous regressors (SARIMAX) model failed to converge for a given window, the forecast defaulted to a naïve persistence estimate (last observed value). This ensured the robustness of the forecasting pipeline while maintaining adaptability to evolving temporal conditions.

### VECM

The VECM extends the VAR framework to handle nonstationary multivariate time series that exhibit cointegration (Shahandashti and Ashuri 2016). By combining short-term differenced dynamics with an error-correction term, VECM captures both temporary fluctuations and long-run equilibrium relationships among variables (Parot et al. 2019). This dual structure makes it particularly suitable for modeling construction cost indices, where prices are influenced by both immediate shocks and persistent economic drivers

$$\Delta z_{i,t} = \alpha_i \beta_i^{\top} z_{i,t-1} + \Gamma_i \Delta z_{i,t-1} + \Psi_i d_t + \varepsilon_{i,t} \tag{20}$$

where $z_{i,t}$ = VECM vector for CSI six-digit Section $i$, containing $y_{i,t}$ and PCA-transformed explanatory components; $\Delta z_{i,t} = z_{i,t} - z_{i,t-1}$; $\alpha_i$ contains adjustment coefficients; $\beta_i$ = cointegration vector; $\Gamma_i$ captures short-run dynamics; $d_t$ contains seasonal indicators or other deterministic terms; $\Psi_i$ = corresponding coefficient matrix; and $\varepsilon_{i,t}$ = VECM error vector.

In this study, each CSI Section index is modeled independently against a reduced set of explanatory variables. Because PPIs are high-dimensional and collinear, they are compressed using principal component analysis (PCA) into the top 10 orthogonal components, which preserve most of the variance while mitigating multicollinearity and avoiding singularity problems in estimation (Abdi and Williams 2010). A quarterly seasonal indicator is also included to account for cyclical effects

$$\hat{y}_{i,t+h|t} = g^{-1}(\hat{\tilde{y}}_{i,t+h|t}) \tag{21}$$

where $\hat{\tilde{y}}_{i,t+h|t}$ = forecast on the transformed scale; and $g^{-1}(\,\cdot\,)$ maps the forecast back to the original price index scale.

### Vanilla Transformer

Transformer networks, originally proposed in Vaswani et al. (2017), are sequence modeling architectures that rely on self-attention mechanisms to capture dependencies across all time steps. Unlike recurrent models, Transformers process sequences in parallel and directly model pairwise interactions between observations, enabling them to effectively learn both short- and long-range temporal relationships. This makes them well suited for forecasting tasks with complex temporal dynamics. Considering the relatively small size of our dataset (78 observations), we first adopt a vanilla Transformer structure to provide a fair and controlled comparison, rather than employing more complex high-capacity variants that may be prone to overfitting in low-data settings (Caetano et al. 2025)

$$X^{\mathrm{seq}}_{i,t} = (s_{i,t-T_{\mathrm{lookback}}+1}, \ldots, s_{i,t}) \tag{22}$$

where $X^{\mathrm{seq}}_{i,t}$ = Transformer input sequence for CSI six-digit Section $i$; $s_{i,t}$ = combined target-and-feature vector defined in Eq. (15); and $T_{\mathrm{lookback}}$ = sequence length

$$Z = X^{\mathrm{seq}}_{i,t} W_e + E^{\mathrm{pos}} \tag{23}$$

where $Z$ = embedded sequence representation; $W_e$ = input-projection matrix; and $E^{\mathrm{pos}}$ = positional encoding

$$Q = ZW_Q, \quad K_{\mathrm{att}} = ZW_K, \quad V = ZW_V \tag{24}$$

where $Q$, $K_{\mathrm{att}}$, and $V$ = query, key, and value matrices; and $W_Q$, $W_K$, and $W_V$ = learnable projection matrices. The key matrix is written as $K_{\mathrm{att}}$ to avoid conflict with other uses of $K$

$$\mathrm{Attention}(Q, K_{\mathrm{att}}, V) = \mathrm{softmax}\left(\frac{QK_{\mathrm{att}}^{\top}}{\sqrt{d_k}}\right)V \tag{25}$$

where $d_k$ = query/key dimension used for scaling; and softmax$(\,\cdot\,)$ converts attention scores into normalized weights

$$Z' = \mathrm{LayerNorm}(Z + \mathrm{Attention}(Q, K_{\mathrm{att}}, V)) \tag{26}$$

where $Z'$ = post-attention representation; LayerNorm$(\,\cdot\,)$ = layer normalization

$$Z'' = \mathrm{LayerNorm}(Z' + \mathrm{FFN}(Z')) \tag{27}$$

where $Z''$ = post-feed-forward representation; and FFN$(\,\cdot\,)$ = position-wise feed-forward network

$$\hat{y}^{\mathrm{TR}}_{i,t+1|t} = W_{\mathrm{out}} Z''_{T_{\mathrm{lookback}}} + b_{\mathrm{out}} \tag{28}$$

where $\hat{y}^{\mathrm{TR}}_{i,t+1|t}$ = one-step-ahead Transformer forecast for CSI six-digit Section $i$; $Z''_{T_{\mathrm{lookback}}}$ = final time-step representation; and $W_{\mathrm{out}}$ and $b_{\mathrm{out}}$ are output-layer parameters.

In this study, the vanilla Transformer architecture consisted of a linear input projection to a 32-dimensional embedding space, followed by sinusoidal positional encoding and a single encoder layer with two attention heads and dropout. A walk-forward evaluation strategy was adopted, where predictions were generated recursively and appended to the input sequence for subsequent forecasts. To enhance adaptability, the model was incrementally fine-tuned after each new observation using all available data up to that point for additional epochs, simulating a continuous learning process in a dynamic forecasting environment.

**Chronos-Bolt**

Chronos-Bolt is a recently developed Transformer-based forecasting model designed to advance state-of-the-art performance in time-series prediction tasks (Ansari et al. 2024). Leveraging self-attention mechanisms, it captures both local and global temporal dependencies without the sequential constraints of recurrent architectures. The architecture of Chronos-Bolt supports parallel sequence processing, improving training efficiency while retaining the capacity to learn long-term dependencies. A key advantage is its probabilistic forecasting capability, which provides both point estimates and uncertainty intervals, critical for real-world applications such as budgeting and procurement (Hornek et al. 2025)

$$\text{Attention}_{\text{CB}}(Q, K_{\text{att}}, V) = \text{softmax}\left(\frac{QK_{\text{att}}^{\top}}{\sqrt{d_k}}\right)V \tag{29}$$

where $\text{Attention}_{\text{CB}}(\cdot)$ = Chronos-Bolt self-attention operation; and the remaining symbols are defined as in Eq. (25).

In this study, Chronos-Bolt was implemented within the AutoGluon framework in three variants to evaluate forecasting performance across CSI six-digit Section datasets. The data were structured into overlapping time-series windows with a lookback horizon of 32 quarters and a forecast horizon of 10 quarters. The first variant was a zero-shot model, directly leveraging pretrained representations without further adaptation. The second variant employed fine-tuning with a fixed budget of 360 s per model, allowing the network to adapt to CSI-specific temporal patterns. The third variant further incorporated a covariate regressor (CAT) with standardized scaling, enabling the model to jointly exploit target dynamics and auxiliary explanatory features

$$\hat{y}^{\text{CB}}_{i,t+h|t} = q^{\text{CB}}_{0.5,i,t+h}, \qquad PI^{1-\alpha}_{i,t+h} = [q^{\text{CB}}_{\alpha/2,i,t+h}, q^{\text{CB}}_{1-\alpha/2,i,t+h}] \tag{30}$$

where $\hat{y}^{\text{CB}}_{i,t+h|t}$ = Chronos-Bolt point forecast for CSI six-digit Section $i$; $q^{\text{CB}}_{p,i,t+h}$ = $p$-quantile forecast; and $\text{PI}^{1-\alpha}_{i,t+h} = (1-\alpha)$-level prediction interval. This quantile form avoids imposing a normal-error assumption.

**Model Training**

All models were trained under a unified protocol to ensure comparability. The dataset was divided into training (85%) and test (15%) sets using a strict chronological split, where the test set comprises the most recent observations.

Within the training data, hyperparameter tuning was conducted using a walk-forward validation scheme tailored for time-series forecasting. Key parameters, including network depth, number of hidden units, lookback, epoch, and regularization terms (dropout and weight decay), were systematically explored. For each candidate hyperparameter configuration, a rolling-origin evaluation was performed, where the model was sequentially trained on past observations and validated on the immediately following period, ensuring strict temporal ordering. To improve robustness, this process was repeated five times using different initial training windows. Model performance was evaluated using the average RMSE across all validation runs, and the hyperparameter combination yielding the lowest average RMSE was selected for the final model. A detailed summary of the hyperparameter search space, tuning configurations, and corresponding results is provided in Appendix II.

To evaluate the role of external drivers, two specifications were implemented. The base model relied only on RSMeans cost indices with engineered time-series features, serving as a benchmark for intrinsic price dynamics. The final model expanded the input space by incorporating PPI series and macroeconomic indicators, capturing broader inflationary and demand-related influences. Comparing these two configurations highlights the contribution of exogenous information to predictive accuracy.

We also trained a unified model across all six-digit CSI Sections alongside the section-specific models trained individually. The unified model shares the same LSTM architecture and training protocol as the section-specific models, ensuring a fair comparison. The only difference lies in the training data: the unified model learns from all sections simultaneously, whereas each section-specific model is trained exclusively on its own section's data. This side-by-side comparison allows us to isolate whether pooling heterogeneous sections improve or degrade forecast accuracy relative to training on each section individually.

Finally, model evaluation on the test set follows a walk-forward incremental procedure. After hyperparameter selection, the final model is retrained on the full training set and deployed sequentially over the test horizon. At each time step, a prediction is first generated for the next period, after which the true observation becomes available and is incorporated into the training set for model updating. This *predict → observe → update → predict next* process ensures that the model is continuously refined using newly available data while strictly preserving temporal causality. At no point does the model access information beyond the prediction timestamp, ensuring a realistic evaluation consistent with real-world deployment. The overall training and evaluation process is illustrated in Fig. 10.

## Results and Discussion

This section presents the comparative results of the proposed forecasting framework and discusses their implications.

### Performance Metrics for Model Evaluations

To assess predictive accuracy, this study employs three standard metrics: the coefficient of determination ($R^2$), RMSE, and mean absolute percentage error (MAPE). Together, these measures provide complementary perspectives on model performance, capturing both explanatory power and error magnitude. $R^2$ quantifies the proportion of variance in the observed data explained by the model. Higher $R^2$ values indicate that the model more effectively reproduces the variability and long-term trends of the time series. RMSE measures the square root of the average squared differences between predicted and observed values. Because it penalizes larger deviations more heavily, RMSE serves as an absolute measure of prediction error and highlights the presence of large forecasting mistakes. Lower values reflect greater overall accuracy. MAPE expresses the average absolute difference between predicted and observed values as a percentage of the observed values. As a scale-independent measure, MAPE facilitates error comparisons across targets of different magnitudes. Lower MAPE values indicate that predictions are, on average, closer to the true values in relative terms.

### Comparisons of Standard Metrics

This subsection compares the performance of five forecasting models, LSTM, ARIMA, VECM, vanilla Transformer and Chronos-Bolt, on CSI six-digit Section price indices within Division 06. Performance is evaluated using $R^2$, RMSE, and MAPE, with results aggregated by both median and mean across all sections to provide robust comparisons. Table 2 reports the primary performance metrics, while Table 3 provides a supplementary paired squared-error loss-differential comparison for the aligned held-out forecasts.

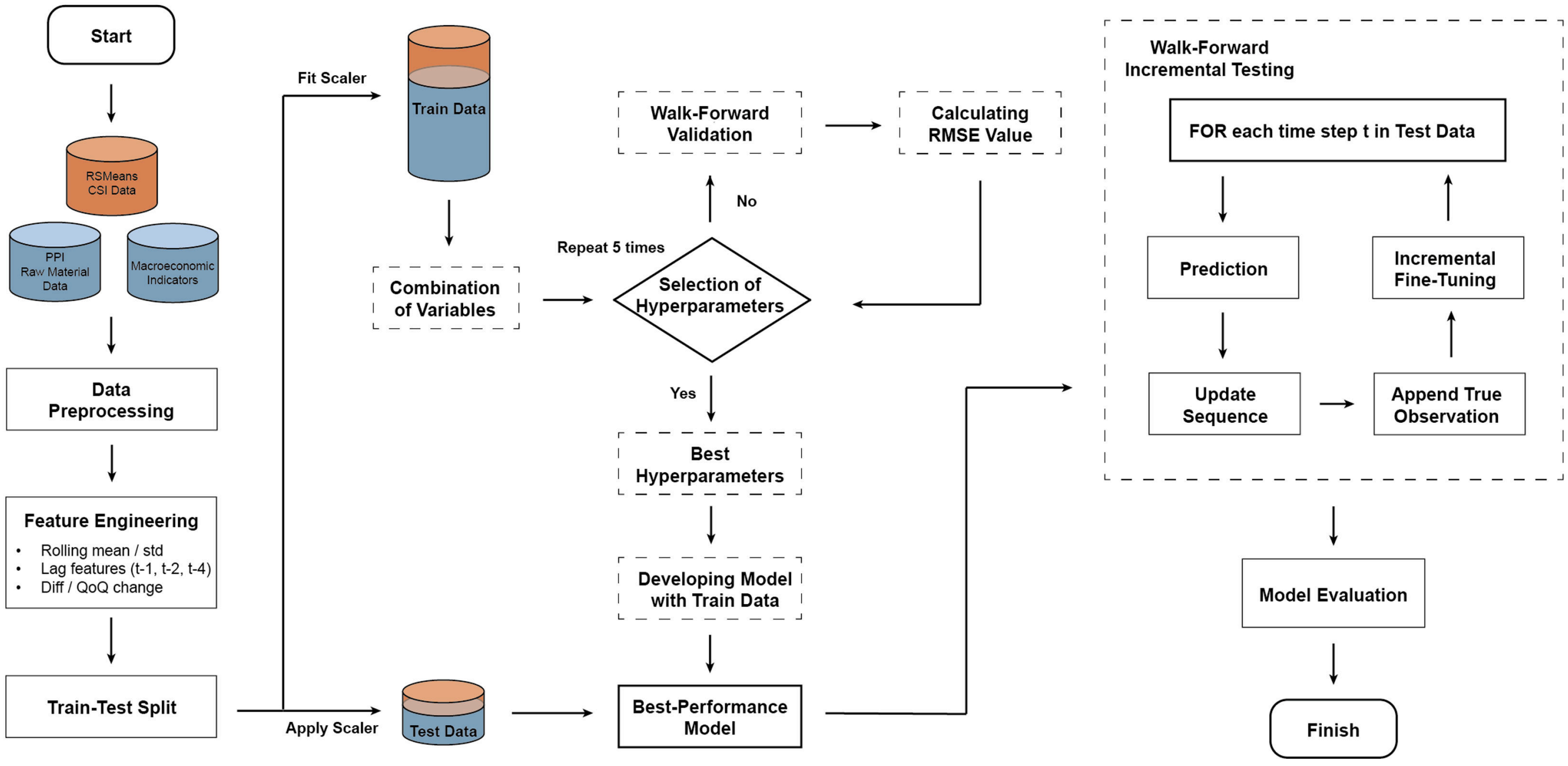

**Fig. 10.** Workflow of the proposed model training, walk-forward validation, and evaluation process.

Table 3 reports supplementary paired loss-differential comparisons using one-step-ahead squared forecast errors over the aligned held-out period. The loss differential is computed as the squared forecast error of the competing model minus the squared forecast error of LSTM; positive values therefore indicate lower squared forecast loss for LSTM. The aligned sample contains 330 Section-quarter paired forecast errors from valid Division 06 six-digit Sections. Section-level $p$-values are computed from Section-average loss differentials, rather than treating all pooled Section-quarter errors as independent observations. Quarter-level and two-way clustered checks were also examined as sensitivity checks for common forecast-quarter shocks and cross-sectional dependence. The comparison is supplementary; primary model evaluation is based on RMSE, MAPE, and $R$-squared in Table 2.

The results in Table 2 reveal a pronounced difference between base and fine-tuned specifications. Base models, trained solely on historical CSI price data, consistently deliver lower accuracy across all metrics. In contrast, fine-tuned models, which incorporate PPI series and macroeconomic indicators, achieve substantially improved forecasts. This highlights the value of including external drivers, which capture broader economic dynamics beyond intrinsic price histories. Aggregating results across Division 06 further clarifies model differences. The LSTM model provides the strongest

**Table 2.** Performance comparison of five forecasting models on Division 06 price indices

| | | Median metric value | | | Mean metric value | | |
|---|---|---|---|---|---|---|---|
| Model | Specification | RMSE | MAPE | $R^2$ | RMSE | MAPE | $R^2$ |
| LSTM | Base | 18.55 | 7.88 | 0.55 | 22.52 | 9.56 | −1.61 |
| | Fine-tuned | **4.19** | **1.60** | **0.96** | **7.47** | **2.36** | **0.92** |
| ARIMA | Base | 32.17 | 16.67 | 0.065 | 38.20 | 15.82 | −4.51 |
| | Fine-tuned | **14.82** | **6.50** | **0.80** | **15.06** | **6.63** | **0.46** |
| VECM | Base | **44.80** | **20.56** | **−1.03** | **54.47** | **23.22** | **−5.56** |
| | Fine-tuned | 53.27 | 25.48 | −2.38 | 71.52 | 28.53 | −7.80 |
| Vanilla Transformer | Base | 37.31 | 17.60 | −0.19 | 36.64 | 16.47 | −0.92 |
| | Fine-tuned | **15.41** | **6.29** | **0.70** | **18.40** | **7.71** | **0.45** |
| Chronos-Bolt | Zero-shot | **48.99** | 25.49 | −1.25 | 58.61 | **25.09** | **−5.15** |
| | Fine-tuned | 55.84 | 25.86 | −1.38 | 58.89 | 25.96 | −10.84 |
| | Fine-tuned + regressor | 49.79 | **22.81** | **−1.04** | **55.18** | 25.12 | −6.43 |

Note: RMSE is reported in the original units of the dependent variable. MAPE is expressed as a percentage, and $R^2$ is unitless. Bold values indicate the best performance within each model group.

**Table 3.** Diebold–Mariano tests of equal predictive accuracy for Division 06 (wood, plastics, and composites)

| Comparison | Mean loss differential, $\bar{d}$ (competing model—LSTM) | $p$-value |
|---|---|---|
| LSTM versus ARIMA | 706.33 | 0.0452 |
| LSTM versus VECM | 2,624.58 | $6.06 \times 10^{-13}$ |

Note: Loss differentials are one-step-ahead ($h = 1$) squared forecast errors, averaged over the aligned held-out sample of 330 Section-quarter pairs; positive $\bar{d}$ indicates lower forecast loss for the LSTM.

performance, with its fine-tuned version achieving a median RMSE of 4.19, a median MAPE of 1.60, and a median $R^2$ of 0.96. Even the base LSTM model outperforms most alternatives, with a median RMSE of 18.55, MAPE of 7.88, and $R^2$ of 0.55. ARIMA also benefits from fine-tuning, improving to a median RMSE of 14.82, MAPE of 6.50, and $R^2$ of 0.80, though it remains less accurate than LSTM. The vanilla Transformer achieves a median RMSE of 15.41, MAPE of 6.29, and $R^2$ of 0.70, reflecting comparable error levels but lower explanatory power. In contrast, VECM and Chronos-Bolt performed poorly. The fine-tuned VECM records a median RMSE of 53.27, MAPE of 25.48, and $R^2$ of −2.38, while the fine-tuned Chronos-Bolt yields a median RMSE of 49.79, MAPE of 22.81, and $R^2$ of −1.04. Despite fine-tuning, both models failed to capture the complex dynamics of construction material prices.

The supplementary loss-differential results in Table 3 are consistent with the RMSE, MAPE, and $R$-squared comparisons in Table 2. For the fine-tuned models, LSTM shows lower mean squared forecast loss than ARIMA and VECM over the aligned held-out sample. The comparison with VECM remains stable under the sensitivity checks, while the ARIMA comparison favors LSTM in mean loss and Section-level inference but is more sensitive to common forecast-quarter effects. Therefore, Table 3 is interpreted as supporting statistical evidence, while the primary model comparison remains based on the performance metrics in Table 2.

Taken together, these aggregate results suggest that LSTM captures richer temporal and nonlinear dependencies than ARIMA and vanilla Transformer, while VECM and Chronos-Bolt appear structurally mismatched to the CSI dataset. Section 4.3 explores these patterns at the Section level.

### CSI 6-Digit Section Price Prediction Results

The Fig. 11 illustrates forecasts generated by the five fine-tuned models for four representative CSI six-digit Sections: 060523 (structural wood framing), 061623 (plastic-based components), 062213 (metal hardware), and 064316 (wood finish carpentry). These sections were selected to capture diverse material categories with distinct market dynamics, including wood, plastics, and metals, thereby providing insight into how models perform under different cost structures. The figures display observed and predicted price indices over Q1 2007–Q2 2025, with the forecast horizon spanning Q1 2023–Q2 2025.

Across all four Sections, the LSTM consistently outperformed alternative approaches, achieving very high $R^2$ values (0.93–0.99), low RMSE (3.30–11.60), and low MAPE (1.32–2.33). The forecasts closely tracked both the long-term trajectory and shorter-term fluctuations, demonstrating LSTM's superior ability to capture nonlinear temporal dependencies and seasonal cycles. By contrast, ARIMA achieved only moderate accuracy, with $R^2$ values of 0.28–0.83, RMSE of 10.59–20.52, and MAPE of 4.78–7.50. While ARIMA effectively captured general upward trends, it systematically lagged during periods of abrupt shifts, such as the rapid inflationary spikes observed in 2021–2022. The vanilla Transformer exhibited a similar performance pattern, with $R^2$ values of 0.45–0.87, RMSE of 9.22–23.29, and MAPE of 4.49–7.36.

VECM and Chronos-Bolt performed poorly across all four Sections. VECM produced negative or weakly positive $R^2$ values (−3.10 to 0.59), high RMSE (13.50–92.52), and large MAPE (6.78–29.96), indicating that forecasts frequently deviated substantially from actual observations. This weakness reflects the model's reliance on linear cointegration assumptions and its sensitivity to dimensionality reduction via PCA, which stripped away important nonlinear dynamics. Similarly, Chronos-Bolt underperformed with $R^2$ values between −1.86 and 0.41, RMSE of 20.74–77.26, and MAPE of 10.65–22.81. Despite its success in broader forecasting domains, Chronos-Bolt's large capacity and reliance on pretrained representations proved unsuitable for this domain-specific dataset, leading to overfitting and unstable predictions.

In summary, Section-level analysis confirms the aggregate results: LSTM achieves stable and accurate forecasts across diverse material categories, ARIMA and vanilla Transformer provide a reasonable but less flexible baseline, and both VECM and Chronos-Bolt fail to adapt to the characteristics of construction price data. Section 4.4 explores the underlying reasons for these performance differences.

We also compared the unified model trained on pooled data from all sections against section-specific models trained individually. Results indicate that section-specific models consistently outperformed the unified alternative. For example, Sections 061623 and 062213, both wood-related, showed substantially higher forecast errors under the unified model. This is because the unified model learns from all sections simultaneously, introducing irrelevant information and high noise from dissimilar sections. Additionally, the unified model incurs considerably higher computational cost. While unified models could provide greater suitability for sparse multiseries corpora by leveraging statistical strength across related sections, the heterogeneity of the present data makes section-specific models more effective. Based on this observation, potential hybrid strategies are discussed in the "Future Work" section.

To further enhance model interpretability and corroborate the findings, SHapley Additive exPlanations (SHAP) analysis was conducted for the LSTM model. This approach enables the quantification of feature contributions to the model's predictions, providing additional insight beyond aggregate performance metrics. The SHAP results confirm that both PPI-related variables and macroeconomic indicators play significant roles in driving material price forecasts, alongside engineered temporal features such as lagged values and rolling statistics. The SHAP visualizations are provided in Appendix III.

SHAP analysis supports decision-making by providing feature-stability analysis, confirming that the relative importance of key drivers remains consistent across multiple random seeds and walk-forward windows. This enhances the credibility of the model's predictions and offers actionable insights for practitioners, identifying which factors most influence a given material section. That said, we do not treat SHAP as a primary evaluation metric. The core assessment of forecast accuracy remains grounded in point-forecast metrics, which are directly comparable across models and widely accepted as the standard in forecasting research. SHAP supplements these metrics as an interpretability tool, offering actionable insights without redefining the evaluation framework.

### Result Analysis

The comparative results reveal clear performance hierarchies and highlight the strengths and weaknesses of each modeling approach.

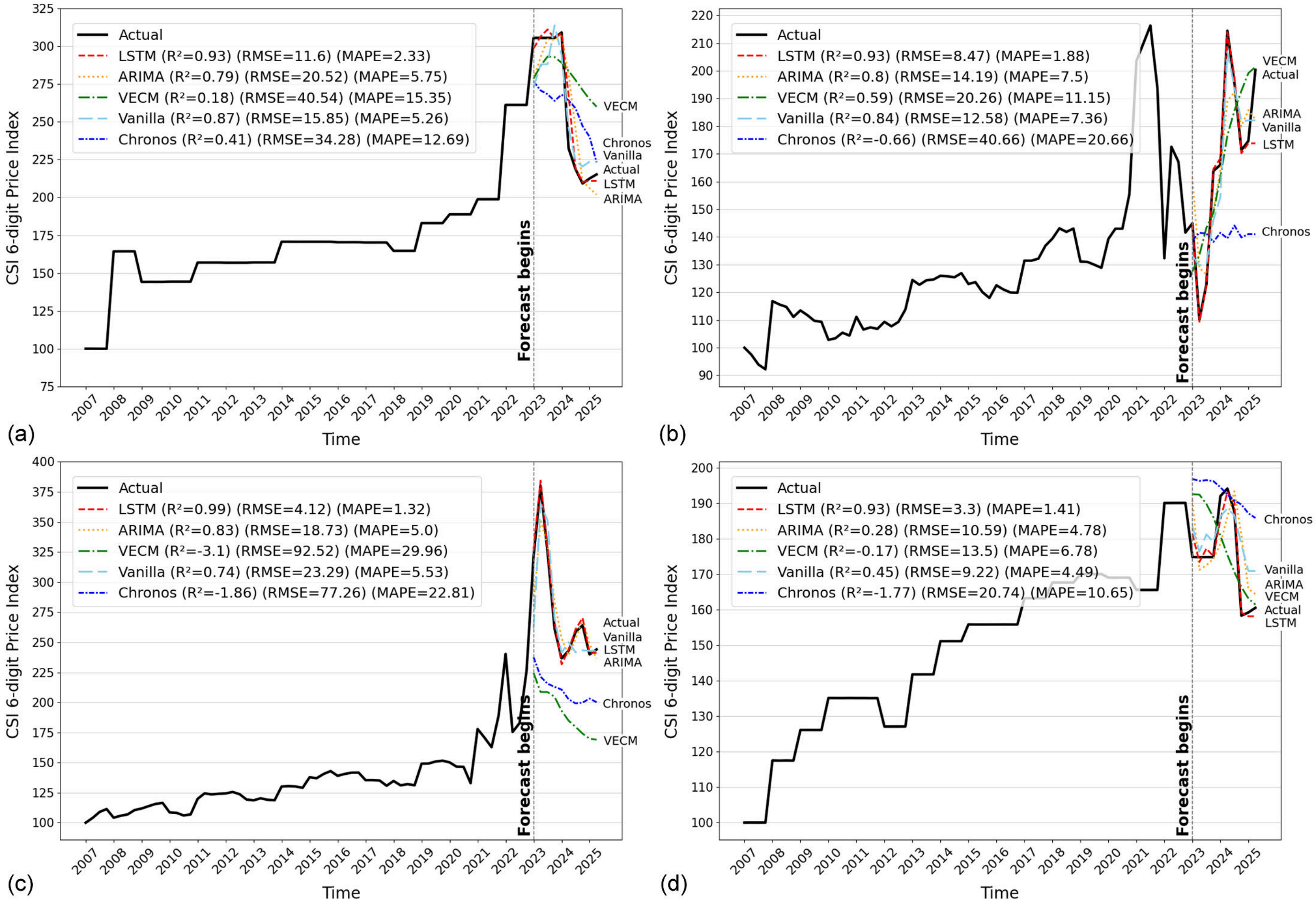

**Fig. 11.** Predicted values of (a) CSI 060523; (b) CSI 061623; (c) CSI 062213; and (d) CSI 064316 Sections using five fine-tuned models.

LSTM outperforms all alternatives due to its ability to capture complex temporal dependencies through gated memory mechanisms. The combination of short-term fluctuations and long-term trends in CSI indices is well suited to LSTM's architecture, which selectively retains relevant information while discarding noise. Its nonlinear modeling capacity also enables effective integration of PPIs and macroeconomic drivers, capturing joint effects that linear methods cannot. Moreover, the incremental fine-tuning strategy used in this study enhances adaptability, allowing the model to update dynamically as new data arrive. These design features explain why LSTM delivers consistently high accuracy and robustness, even under volatile market conditions.

ARIMA, while less flexible, performs reasonably well because of its strength in modeling autoregressive dynamics and seasonal patterns. The rolling SARIMAX implementation ensures responsiveness to local shifts, while exogenous predictors improve explanatory power. However, ARIMA is inherently limited to linear dependencies over short horizons and cannot capture nonlinear interactions or structural breaks. This explains why it performs acceptably but lags behind LSTM in accuracy.

The vanilla Transformer, while more flexible than traditional econometric models, demonstrates reasonably competitive performance due to its ability to model long-range dependencies through self-attention mechanisms. By assigning adaptive weights to historical observations, it can capture temporal relationships that are not restricted by fixed recurrence structures. However, the vanilla Transformer remains limited in its ability to learn effectively under small-sample conditions, as its performance heavily depends on sufficient data to stabilize attention patterns. As a result, it performs better than high-capacity pretrained Transformer models such as Chronos-Bolt, which tend to overfit in data-scarce settings, but still underperforms compared to LSTM, which is better suited for capturing structured temporal dynamics with limited data.

VECM struggles primarily due to structural incompatibility with the dataset. The VECM framework is based on a linear cointegration formulation, which can be expressed as:

$$\Delta z_{i,t} = \Pi_i z_{i,t-1} + \sum_{a=1}^{p_V} \Gamma_{i,a} \Delta z_{i,t-a} + \Psi_i d_t + \varepsilon_{i,t} \tag{31}$$

where $z_{i,t}$ = local VECM variable vector for CSI six-digit Section $i$; $\Pi_i$ captures long-run equilibrium relationships; $\Gamma_{i,a}$ captures short-run dynamics at differenced lag $a$; $p_V$ = number of differenced lags in the VECM; $d_t$ contains deterministic terms; and $\Psi_i$ = corresponding coefficient matrix.

In this study, the input space consists of over 200 variables, including multiple PPI series and macroeconomic indicators. Under such conditions, the VECM becomes fundamentally difficult to estimate. Specifically, the dimensionality of the parameter matrices (e.g., $\Pi_i$ and $\Gamma_i$) increases quadratically with the number of variables, leading to severe overparameterization and singular or near-singular covariance matrices. To partially address this issue, PCA was applied to reduce dimensionality. However, this transformation compresses the original feature space into a smaller set of orthogonal components, which inevitably removes idiosyncratic signals and nonlinear interactions embedded in the original variables. Consequently, while PCA improves numerical feasibility, it further limits the model's ability to capture the underlying data-generating process.

Chronos-Bolt underperforms despite its strong reputation in broader time-series forecasting. Its failure stems from two main issues: (1) the CSI dataset's relatively short history (≈78 quarterly observations) is insufficient for training such a high-capacity transformer, leading to overfitting; and (2) its reliance on broad pretraining makes it less effective at incorporating structured domain-specific covariates such as PPIs. While Chronos-Bolt excels in domains with large-scale, unstructured series, its architecture does not translate well to this task without extensive fine-tuning and domain-specific adaptation.

Overall, the analyses confirm that LSTM is the most effective approach for CSI price forecasting, combining interpretability of temporal structure with adaptability to nonlinear and heterogeneous dynamics. ARIMA and vanilla Transformer provide a serviceable benchmark but lack flexibility. The insufficient performance of VECM and Chronos-Bolt is framed as evidence of methodological boundaries under different data regimes. These outcomes reflect the inherent suitability of each modeling approach given the characteristics of the dataset.

### *Limitations and Applications*

As further validation, the proposed framework was tested across 22 CSI Divisions at the six-digit level. While performance is strong overall, limitations must be acknowledged. Fig. 12 illustrates results within Division 06, where the framework achieves median values of RMSE = 4.19 and $R^2 = 0.96$, confirming reliable forecasting accuracy. However, certain sections, such as 064439, 065113, and 066310, exhibited relatively weaker performance compared to others, with RMSE values as high as 27 and $R^2$ as low as 0.55.

Detailed data analysis suggests three main reasons for these weaker results. First, although obvious source-data errors and isolated non-market spikes were reviewed and corrected during preprocessing where appropriate, some subsections still exhibit residual mixed-unit inconsistencies and abrupt anomalous movements, which introduce noise into Section-level forecasting. Second, product heterogeneity within some sections (e.g., cedar posts, engineered wood beams, and large solid columns) creates wide and shifting variance that challenges a single model, especially with short historical samples. Finally, the present study is intentionally focused on US residential construction materials, with Division 06 used as the primary demonstration because of its strong relevance to residential construction. Results may limit generalizability across other divisions, commercial construction, or international contexts.

Despite these limitations, the framework demonstrates clear practical value. The primary application of the proposed methodology lies in construction cost estimation, where accurate and timely forecasting of material costs is essential for project planning and financial control (Nguyen et al. 2018). Among the established estimate types, the framework is most appropriately applied at the engineer's and definitive estimate levels for the developer and contractor. At these stages, sufficiently detailed cost information is required to support design development, evaluate material alternatives, and refine project budgets. The framework's ability to generate CSI six-digit Section-level forecasts aligns with this level of detail and can therefore improve the reliability of escalation assumptions used in preconstruction planning and cost review. In this sense, the reported forecasting gains support the material input-cost portion of definitive estimating through more reliable CSI six-digit unit-cost updates and escalation assumptions, rather than by directly measuring final bid or total project cost accuracy.

While optimized for definitive estimating, the framework can also serve as a supplementary reference in other stages of cost estimation. Its high granularity may exceed the typical requirements of conceptual and preliminary estimates; however, this study is intentionally designed to forecast pre-markup construction input costs at the CSI six-digit level. Accordingly, contractor overhead, profit, contingency, and competitive bid effects are excluded by design so that the model isolates raw input-price signals. The framework is therefore most directly suited to engineer's and definitive estimating and support developer decision-making, whereas direct forecasting of bid-level final prices would require an additional markup and market-competition layer beyond the scope of the present study.

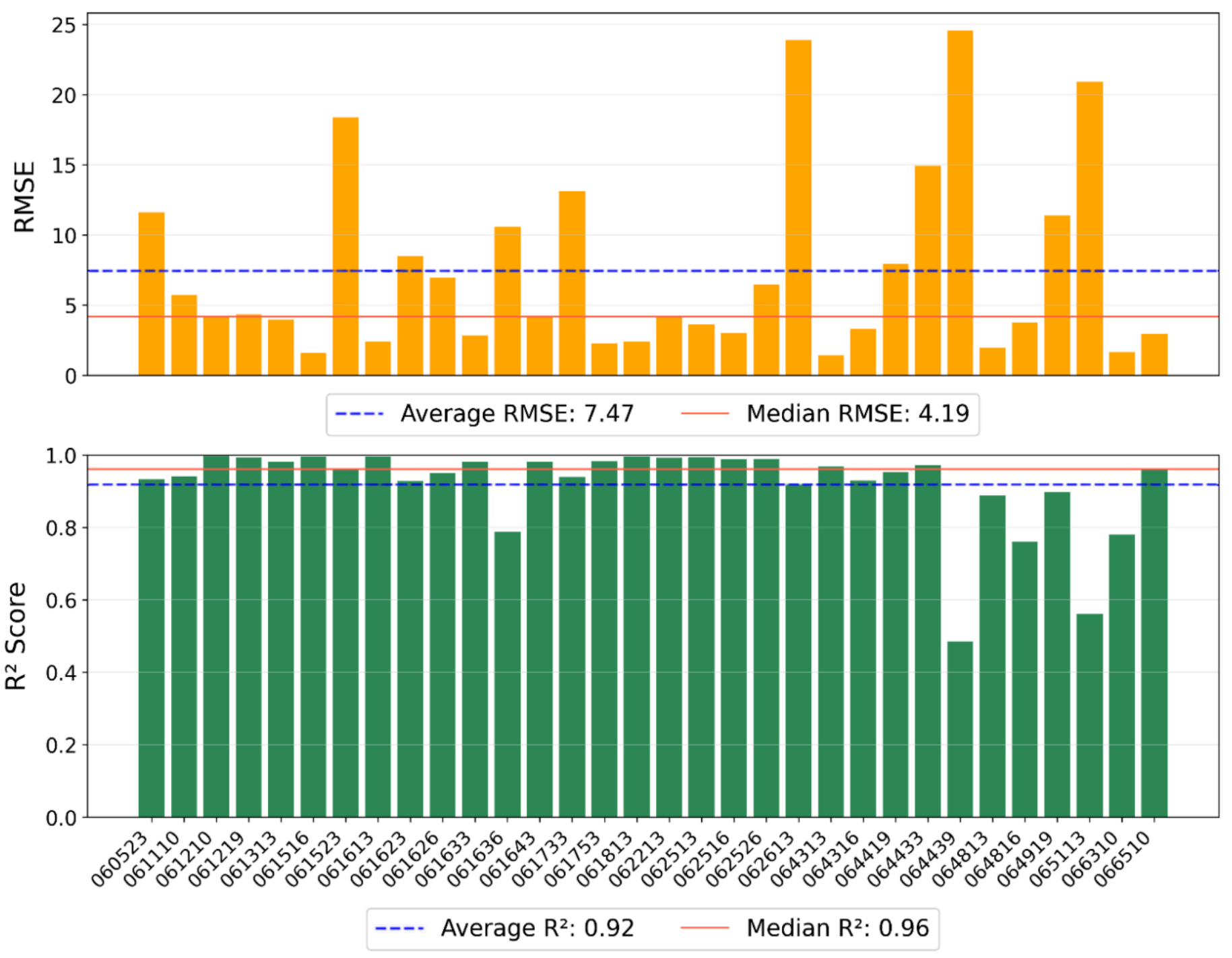

**Fig. 12.** Model performance of fine-tuned LSTM across all 6-digit CSI Sections in Division 06.

Beyond conventional estimating workflows, the same section-level forecasts can be translated into input-cost scenarios for project finance models, real estate investment trust (REIT) underwriting, and commodity hedging analyses, tightening the link between engineering cost estimates and financial risk assessment.

## Conclusion

This study developed and evaluated a scalable forecasting framework for construction material prices that makes three main contributions to construction cost management research. First, it introduces a CSI six-digit forecasting corpus that enables Section-level analysis across multiple material categories. Second, it provides a benchmark across econometric and machine-learning approaches under consistent training and evaluation protocols. Third, it demonstrates how Section-level material price forecasting can support preconstruction decision-making by improving unit-cost updates, procurement planning, and budget review.

Among the five approaches tested, LSTM, ARIMA, VECM, vanilla Transformer, and Chronos-Bolt, the proposed framework with LSTM consistently achieved the highest predictive accuracy. Fine-tuned LSTM models, incorporating external indicators, outperformed both baseline specifications and competing models, with a median $R^2$ of 0.96 and RMSE of 4.19 across Division 06. These results underscore the advantage of deep recurrent architectures in capturing nonlinear temporal dependencies and heterogeneous material dynamics. In contrast, ARIMA provided a reasonable benchmark but was limited by linear assumptions, and the vanilla Transformer achieved competitive performance but was constrained by the limited sample size, while VECM and Chronos-Bolt struggled with dimensionality and data adaptation, respectively.

The framework offers practical applications in budgeting, procurement, and project planning. Accurate Section-level forecasts can help contractors and developers anticipate material price volatility, support public agencies in managing infrastructure budgets, and enable suppliers to adjust production to expected demand. By embedding predictive models into cost estimation workflows, this approach enhances transparency, improves decision-making, and reduces financial risks in construction projects.

Future research will address current limitations of the study discussed in Section 4.5. Efforts will focus on resolving inconsistencies in CSI Section data, detecting anomalies such as abrupt price jumps, and refining forecasts for weaker material categories (e.g., 064439). Testing models at the item level with hierarchical reconciliation to 6-digit and division totals will allow more precise analysis while preserving aggregation consistency. In addition, further research could pair these granular material price forecasts with firm- and asset-level data opens a path to studying how construction input-cost shocks are priced in equity, debt, REIT, and housing markets, and how they shape investment timing, leverage, and risk premia. This research could pair these granular material price forecasts with firm- and asset-level data, which opens a path to studying how construction input-cost shocks are priced in equity, debt, REIT, and housing markets, and how they shape investment timing, leverage, and risk premia. Future work also could explore hybrid strategies that balance the trade-off between section-specific and unified modeling. One promising direction is to cluster sections with strong cross-correlations and shared underlying drivers, then train a unified model per cluster. This approach would retain the specialization benefits of section-specific models while also obtaining information sharing advantages of unified models.

## Appendix I. Training–Validation Loss Curves and Learning Curves

The training–validation loss curves and learning curves are presented in Fig. 13 to evaluate the generalization behavior of the LSTM model and validate the selected hyperparameter configuration. The loss curves illustrate the evolution of model fitting over training epochs, where close alignment between training and validation losses indicates stable learning without overfitting. The learning curves further examine model performance under different training set sizes, where a consistent decrease in both training and validation errors without a widening gap suggests low variance and good generalization capability.

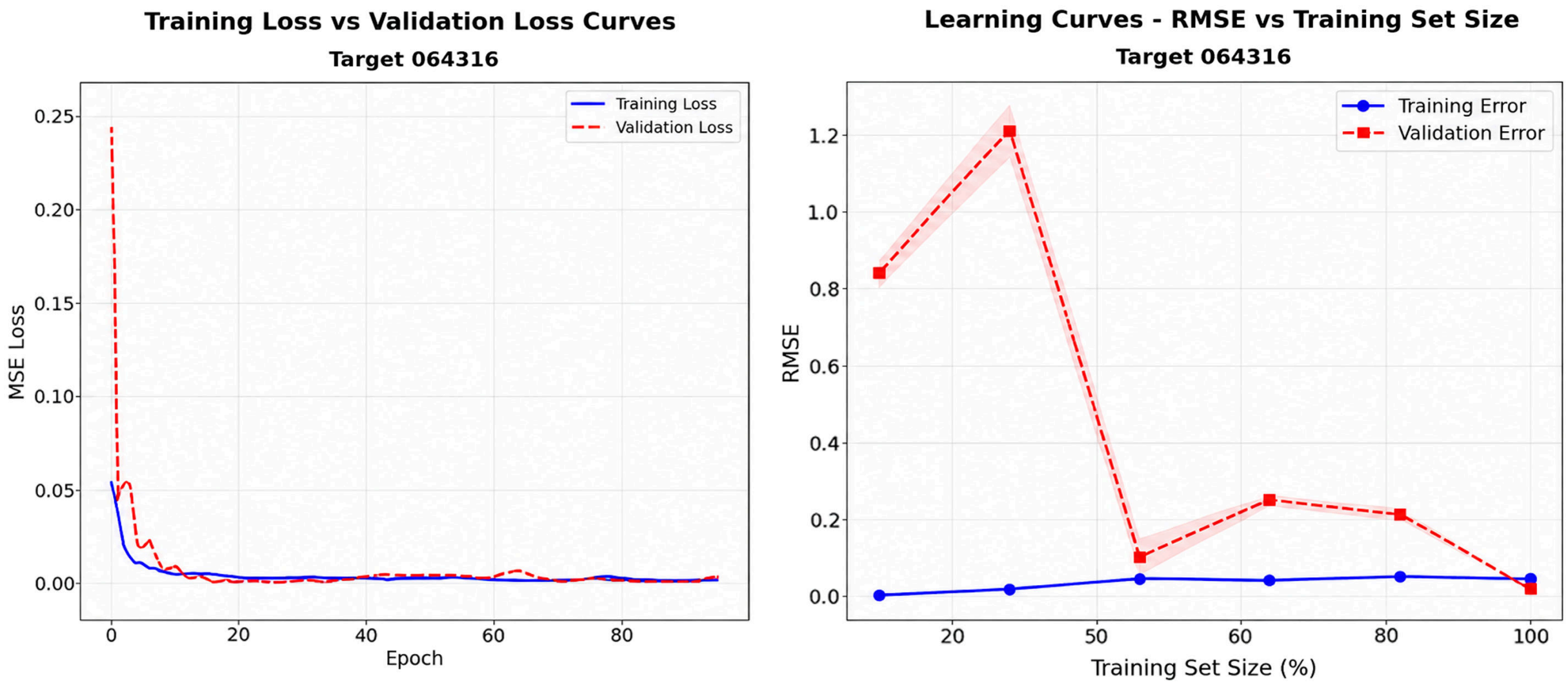

**Fig. 13.** Training and validation loss and learning curves for CSI Section 06 43 16.

These results collectively support the appropriateness of the chosen LSTM architecture for the given dataset and confirm that the model does not exhibit significant overfitting despite its relatively high capacity. Section 064316 is used as a representative example.

## Appendix II. Hyperparameter Tuning Results

The hyperparameter tuning results corresponding to the model development process described in the "Model Training" section are summarized in Table 4. Table 4 presents a comparison of representative hyperparameter configurations explored in this study, focusing on key hyperparameters including network depth, number of hidden units, lookback window, training epochs, dropout rate, and weight decay, among which the selected baseline configuration is identified.

Each table reports the model performance evaluated using walk-forward validation, where results are expressed in terms of the average RMSE across multiple rolling-origin runs. These tables collectively illustrate the impact of individual hyperparameters on model performance and support the selection of the final configuration reported in the main text.

## Appendix III. SHAP Analysis

The SHAP visualizations provide a detailed interpretation of the LSTM model by quantifying the contribution of each input feature to the predicted material prices. In the SHAP summary plot, features are ranked by their overall importance based on the mean absolute SHAP values. Each point represents an individual observation, with the horizontal position indicating the magnitude and direction of the feature's impact on the prediction. The open, half-filled, and filled circles denote the feature value, allowing for the identification of how feature levels influence model outputs.

These visualizations complement the quantitative evaluation in the main text by offering insight into the underlying drivers of the model's predictions. Section 060523 is presented as a representative example, while similar patterns are observed across other sections. Fig. 14 presents the SHAP summary of feature contributions for Section 060523, and Fig. 15 illustrates the top 15 features ranked by mean absolute SHAP values.

**Table 4.** Comparison of representative hyperparameter configurations for the LSTM model

| Hidden units | Layers | Dropout | Epochs | Fine-tune epochs | Lookback | RMSE |
|---|---|---|---|---|---|---|
| 256 | 1 | 0.02 | 100 | 50 | 8 | 5.6 |
| **512** | **1** | **0.02** | **100** | **50** | **8** | **4.19** |
| 1,024 | 1 | 0.02 | 100 | 50 | 8 | 4.81 |
| 512 | 2 | 0.02 | 100 | 50 | 8 | 54.92 |
| 512 | 1 | 0.1 | 100 | 50 | 8 | 7.16 |
| 512 | 1 | 0.02 | 150 | 50 | 8 | 4.92 |
| 512 | 1 | 0.02 | 100 | 80 | 8 | 4.61 |
| 512 | 1 | 0.02 | 100 | 50 | 12 | 5.04 |

Note: The best-performing configuration is highlighted in bold.

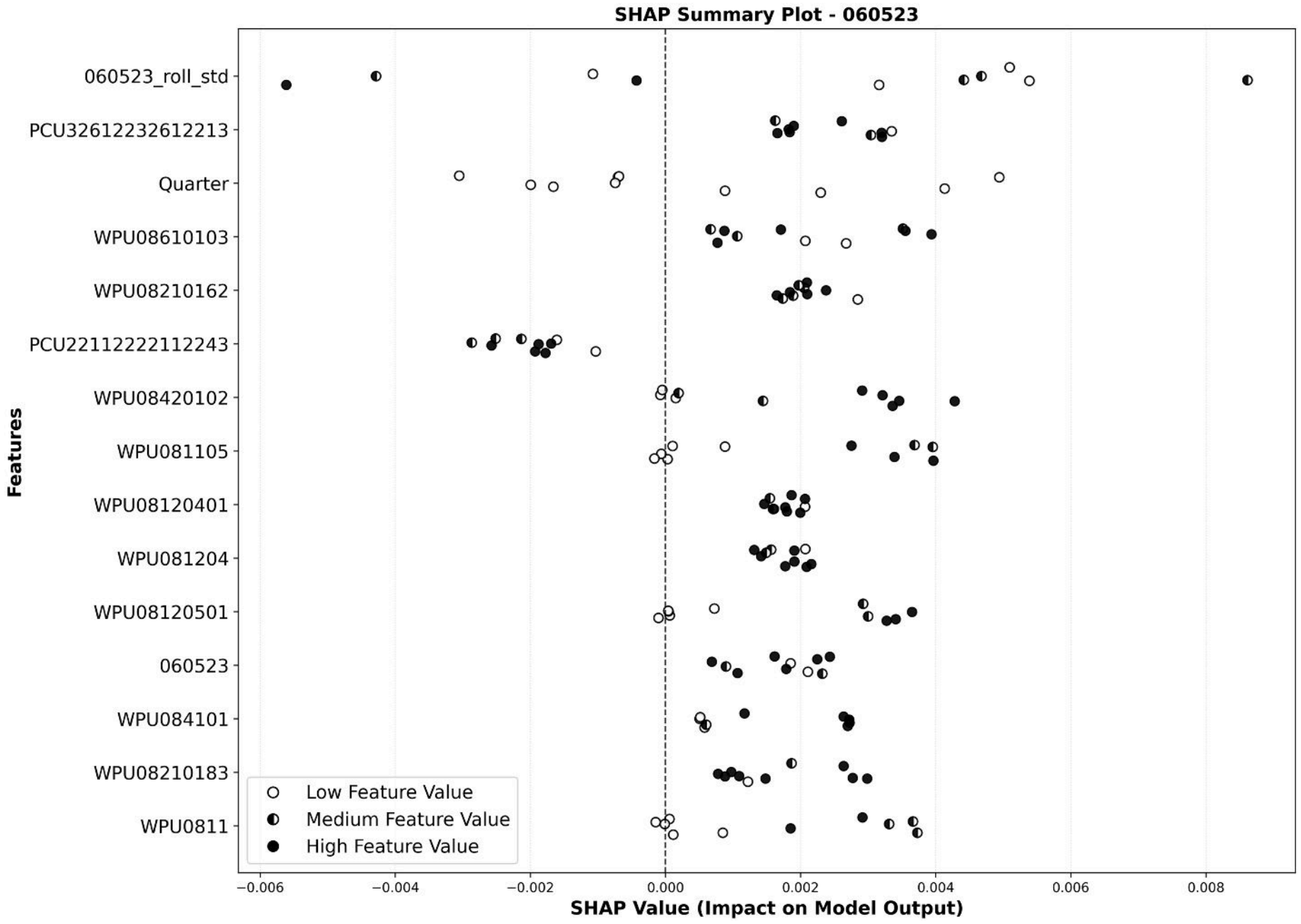

**Fig. 14.** SHAP summary plot of feature contributions for CSI Section 06 05 23.

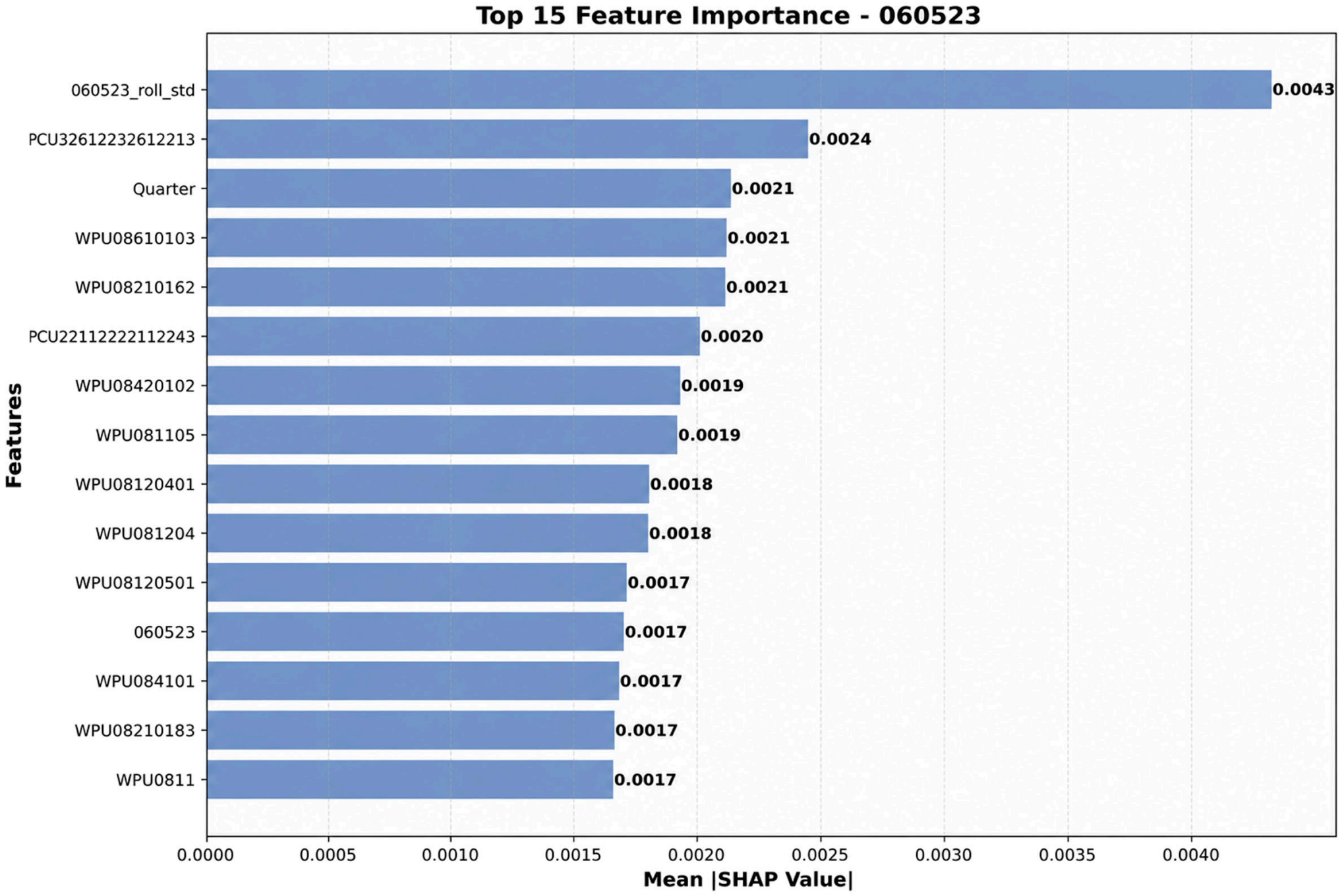

**Fig. 15.** SHAP Top 15 SHAP Feature Importance for CSI Section 06 05 23.

## Data Availability Statement

The data used in this study come from a combination of proprietary and publicly available sources. Construction material price data were obtained from RSMeans Construction Cost Data (Gordian), which are subject to licensing restrictions and therefore cannot be publicly shared. Publicly available explanatory data, including PPI series, CPI, employment statistics, gross domestic product, housing starts, and other macroeconomic indicators, were obtained from the US BLS, the US BEA, and the US Census Bureau. These data are freely accessible from their respective official websites. All preprocessing procedures, feature engineering steps, and model specifications are fully described in the paper. Derived and aggregated datasets, along with representative scripts used for data processing and model implementation (excluding proprietary RSMeans inputs), may be made available by the corresponding author upon reasonable request for academic and noncommercial research purposes.

## Acknowledgments

This research was conducted by the authors as part of internal research and development activities at Quotr AI (https://quotr.ai/). The analyses and conclusions presented in this paper reflect the independent technical judgment of the authors. The authors gratefully acknowledge Ke Ma, a student in the UC Berkeley College of Computing, Data Science and Society and an Applied Machine Learning Engineer Intern at Quotr AI, for support in streamlining the test dataset, investigating additional models, and enhancing the visualization of results during the revision stage.

## Author Contributions

Boge Lyu: Conceptualization; Data curation; Formal analysis; Investigation; Methodology; Software; Validation; Visualization; Writing – original draft; Writing – review and editing. Qianye Yin: Conceptualization; Data curation; Formal analysis; Investigation; Methodology; Validation; Visualization; Writing – original draft; Writing – review and editing. Iris Denise Tommelein: Conceptualization; Investigation; Supervision; Visualization; Writing – review and editing. Hanyang Liu: Conceptualization; Funding acquisition; Resources. Karnamohit Ranka: Formal analysis; Methodology; Software; Validation. Karthik Yeluripati: Software. Junzhe Shi: Conceptualization; Funding acquisition; Resources; Supervision; Writing – review and editing.

## Notation

*The following symbols are used in this paper:*

$a$ = index of differenced lags in the VECM expression;
$B$ = backshift operator, so $By_{i,t} = y_{i,t-1}$;
$b_{\cdot}$ = learnable bias term in neural-network equations;
$c_r$ = LSTM cell state at recurrent step $r$;
$\tilde{c}_r$ = candidate LSTM cell state at recurrent step $r$;
$D$ = seasonal differencing order in the seasonal ARIMAX model;
$D_{q,t}$ = seasonal dummy variable for calendar quarter $q$ at time $t$;
$d$ = nonseasonal differencing order in the ARIMAX model;
$d_k$ = query/key dimension used in scaled dot-product attention;
$d_t$ = vector of deterministic terms, such as seasonal indicators, at quarter $t$;
$E^{\text{pos}}$ = positional encoding matrix in the Transformer model;
$f(\cdot)$ = forecasting function;
$f_h^{(m)}(\cdot)$ = forecasting function for model $m$ and forecast horizon $h$;
$f_r$ = forget gate in the LSTM model at recurrent step $r$;
$g(\cdot)$ = transformation function applied to the target variable;

$g^{-1}(\cdot)$ = inverse transformation function;
$H$ = maximum forecast horizon;
$h$ = forecast horizon;
$h_r$ = LSTM hidden state at recurrent step $r$;
$i$ = index of CSI six-digit Sections;
$J$ = number of explanatory variables;
$j$ = index of explanatory variables;
$K_{\text{att}}$ = key matrix in the self-attention mechanism;
$k$ = rolling-window length;
$L$ = lookback length;
$MA_{i,t}^{(k)}$ = moving average of the material price index for CSI six-digit Section $i$ over a $k$-quarter window ending at quarter $t$;
$m$ = forecasting model;
$N$ = number of CSI six-digit Sections;
$n$ = number of held-out forecast observations used in metric calculations;
$o_r$ = output gate in the LSTM model at recurrent step $r$;
$P$ = seasonal autoregressive order in the seasonal ARIMAX model;
$PI_{i,t+h}^{1-\alpha}$ = $(1-\alpha)$-level prediction interval for CSI six-digit Section $i$ at quarter $t+h$;
$p$ = nonseasonal autoregressive order in the ARIMAX model;
$p_V$ = number of differenced lags in the VECM;
$Q$ = query matrix in the self-attention mechanism;
$Q(t)$ = calendar-quarter function that returns the quarter associated with time $t$;
$q$ = nonseasonal moving-average order in the ARIMAX model, or calendar-quarter index when used in $D_{q,t}$;
$q_{p,i,t+h}^{CB}$ = $p$-quantile Chronos-Bolt forecast for CSI six-digit Section $i$ at quarter $t+h$;
$R^2$ = coefficient of determination used to evaluate forecast accuracy;
$r$ = local recurrent time-step index in the LSTM model, or held-out forecast-observation index in metric equations;
$SD_{i,t}^{(k)}$ = rolling standard deviation of the material price index for CSI six-digit Section $i$ over a $k$-quarter window ending at quarter $t$;
$s$ = seasonal period in the ARIMAX model;
$s_{i,t}$ = combined target-and-feature input vector for CSI six-digit Section $i$ at quarter $t$;
$T$ = number of quarters in the full time series;
$T_{\text{lookback}}$ = sequence length used by the LSTM, Transformer, and Chronos-Bolt models;
$t$ = quarterly time index;
$V$ = value matrix in the self-attention mechanism;
$W_{\cdot}$ = learnable weight matrix in neural-network equations;
$X_t$ = vector of explanatory variables available at quarter $t$;
$X_{i,t}^{\text{seq}}$ = input sequence for CSI six-digit Section $i$ ending at quarter $t$;
$x_{j,t}$ = explanatory variable $j$ at quarter $t$;
$x_{j,t}^{\text{CCI}}$ = city-level cost-index explanatory variable $j$ at quarter $t$;
$x_{j,t}^{\text{macro}}$ = macroeconomic explanatory variable $j$ at quarter $t$;
$x_{j,t}^{\text{PPI}}$ = PPI-based explanatory variable $j$ at quarter $t$;
$Y_t$ = vector of material price indices for all CSI six-digit Sections at quarter $t$;
$y_{i,t}$ = rebased material price index for CSI six-digit Section $i$ at quarter $t$;
$y_r$ = observed value for held-out forecast observation $r$;
$\hat{y}_{i,t+h|t}^{(m)}$ = forecast of $y_{i,t+h}$ for CSI six-digit Section $i$, made at quarter $t$ using model $m$;
$\hat{y}_r$ = forecast corresponding to held-out observation $r$;
$\tilde{y}_{i,t}$ = transformed target value for CSI six-digit Section $i$ at quarter $t$;
$\bar{y}$ = mean of observed values in the evaluation sample;
$Z$ = embedded sequence representation in the Transformer model;
$Z'$ = post-attention sequence representation in the Transformer model;
$Z''$ = post-feed-forward sequence representation in the Transformer model;
$z_{i,t}$ = VECM variable vector for CSI six-digit Section $i$, containing the target value and reduced explanatory components;
$\alpha$ = significance level for prediction intervals;
$\alpha_i$ = VECM adjustment-coefficient vector for CSI six-digit Section $i$;
$\beta_i$ = VECM cointegration vector for CSI six-digit Section $i$;
$\Gamma_i$ = VECM short-run coefficient matrix for CSI six-digit Section $i$;
$\Gamma_{i,a}$ = VECM short-run coefficient matrix for CSI six-digit Section $i$ at differenced lag $a$;
$\Delta z_{i,t}$ = first difference of the VECM vector $z_{i,t}$;
$\Delta_4 y_{i,t}$ = four-quarter change in the material price index for CSI six-digit Section $i$;
$\delta$ = coefficient vector for exogenous variables in the ARIMAX model;
$\varepsilon_{i,t}$ = model error term for CSI six-digit Section $i$ at quarter $t$;
$\varepsilon_{i,t+h}^{(m)}$ = forecast error for CSI six-digit Section $i$, forecast horizon $h$, and model $m$;
$\Pi_i$ = VECM long-run coefficient matrix for CSI six-digit Section $i$;
$\Theta_Q(B^s)$ = seasonal moving-average polynomial in the ARIMAX model;
$\theta_i^{(m)}$ = parameters of model $m$ estimated for CSI six-digit Section $i$;
$\theta_q(B)$ = nonseasonal moving-average polynomial in the ARIMAX model;
$\iota_r$ = input gate in the LSTM model at recurrent step $r$;
$\sigma(\cdot)$ = sigmoid activation function;
$\tau$ = lag index;
$\Phi_P(B^s)$ = seasonal autoregressive polynomial in the ARIMAX model;
$\phi_p(B)$ = nonseasonal autoregressive polynomial in the ARIMAX model;
$\Psi_i$ = VECM coefficient matrix for deterministic terms for CSI six-digit Section $i$;
$\mathbb{1}\{\cdot\}$ = indicator function equal to 1 when the condition is true and 0 otherwise;
$\mathbb{E}[\cdot\mid\cdot]$ = conditional expectation operator;
$\mathcal{I}_{i,t}^{\text{base}}(L)$ = base information set for CSI six-digit Section $i$ at quarter $t$; and
$\mathcal{I}_{i,t}^{\text{aug}}(L)$ = augmented information set for CSI six-digit Section $i$ at quarter $t$.